\documentclass{article}

\usepackage[preprint, nonatbib]{neurips_2021}

\usepackage[utf8]{inputenc} 
\usepackage[T1]{fontenc}    
\usepackage{hyperref}       
\usepackage{url}            
\usepackage{booktabs}       
\usepackage{amsfonts}       
\usepackage{nicefrac}       
\usepackage{microtype}      
\usepackage[table]{xcolor}         

\usepackage{amsmath}
\usepackage[title]{appendix}
\usepackage[utf8]{inputenc}
\usepackage{lipsum}                     
\usepackage{xargs}                      
\usepackage[colorinlistoftodos,prependcaption,textsize=tiny]{todonotes}
\newcommandx{\unsure}[2][1=]{\todo[linecolor=red,backgroundcolor=red!25,bordercolor=red,#1]{#2}}
\newcommandx{\change}[2][1=]{\todo[linecolor=blue,backgroundcolor=blue!25,bordercolor=blue,#1]{#2}}
\newcommandx{\info}[2][1=]{\todo[linecolor=OliveGreen,backgroundcolor=OliveGreen!25,bordercolor=OliveGreen,#1]{#2}}
\newcommandx{\improvement}[2][1=]{\todo[linecolor=Plum,backgroundcolor=Plum!25,bordercolor=Plum,#1]{#2}}
\newcommandx{\thiswillnotshow}[2][1=]{\todo[disable,#1]{#2}}
\usepackage{titlesec}
\newcommand{\m}{\mathbb}

\usepackage[utf8]{inputenc}
\usepackage{multirow}
\usepackage{amssymb}
\usepackage{graphicx}
\usepackage[nottoc]{tocbibind}
\usepackage{blindtext}

\usepackage{diagbox}
\usepackage{booktabs, multirow} 
\usepackage{soul}
\usepackage{changepage,threeparttable} 
\usepackage{tabularx}
\usepackage{colortbl}
\usepackage{tablestyles}

\usepackage[symbol,hang,flushmargin]{footmisc}

\title{SMPL: Simulated Industrial Manufacturing and Process Control Learning Environments}

\author{%
  {Mohan Zhang$^{1,2}$, \ Xiaozhou Wang$^{1}$, \ Benjamin Decardi-Nelson$^{3}$, \ Song Bo$^{3}$}, \\
  {\textbf{An Zhang$^{3}$, \ Jinfeng Liu$^{3}$, \ Sile Tao$^{1}$, \ Jiayi Cheng$^{1}$, \ Xiaohong Liu$^{5}$}}, \\
  {\textbf{DengDeng Yu$^{4}$, \ Matthew Poon$^{1}$, \ Animesh Garg$^{2}$}}\\
  $^1$Quartic.ai \\
  \texttt{\small mohan, xiaozhou, bill.tao, jerry, matthew@quartic.ai}\\
  $^2$University of Toronto \\
  \texttt{\small zhangmo4, garg@cs.toronto.edu} \\
  $^3$University of Alberta \\
  \texttt{\small decardin, sbo, azhang1, jinfeng@ualberta.ca} \\
  $^4$University of Texas at Arlington \\
  \texttt{\small dengdeng.yu@uta.edu} \\
  $^5$Shanghai Jiao Tong University \\
  \texttt{\small xiaohongliu@sjtu.edu.cn}
  }

\begin{document}

\maketitle

\begin{abstract}
  Traditional biological and pharmaceutical manufacturing plants are controlled by human workers or pre-defined thresholds. Modernized factories have advanced process control algorithms such as model predictive control (MPC). However, there is little exploration of applying deep reinforcement learning to control manufacturing plants. One of the reasons is the lack of high fidelity simulations and standard APIs for benchmarking. To bridge this gap, we develop an easy-to-use library that includes five high-fidelity simulation environments: BeerFMTEnv, ReactorEnv, AtropineEnv, PenSimEnv and mAbEnv, which cover a wide range of manufacturing processes. We build these environments on published dynamics models. Furthermore, we benchmark online and offline, model-based and model-free reinforcement learning algorithms for comparisons of follow-up research.
  \footnote[2]{Official documentation:\href{https://smpl-env.github.io/smpl-document/index.html}{https://smpl-env.github.io/smpl-document/index.html}\\
  Official implementation: \href{https://github.com/smpl-env/smpl}{https://github.com/smpl-env/smpl}\\
  Code of experiments: \href{https://github.com/smpl-env/smpl-experiments}{https://github.com/smpl-env/smpl-experiments}\\
  
  }

\end{abstract}

\section{Introduction}
With a large market value~\cite{thomas2020annual}, the manufacturing industry is enthusiastic to the ways that is conductive to the production efficiency.
A number of studies have demonstrated that reinforcement learning may be applied to manufacturing processes and has the potential to dramatically improve productivity \cite{mahadevan1998optimizing, OVERBECK2021170}. 

Our goal is to bridge the gap between deep reinforcement learning research and industrial manufacturing by creating simulation environments that model real-world factories. In this paper, we introduce five manufacturing simulation environments, including beer fermentation, atropine production, penicillin manufacturing, monoclonal antibodies production, and a continuous stirred tank simulation. These simulated environments allow us to test the latest advances in reinforcement learning in controlled environments without safety concerns.


In reinforcement learning, the environment is commonly modeled as a Markov Decision Process (MDP). In SMPL, the \textbf{state space} is defined by the collection of all reactions, material flows, their concentrations and the state of where the reactions are taking place (e.g. the temperature of a reactor tank). The \textbf{initial state} of a trajectory is randomly sampled within a reasonable range, which involves stochasticity. For the \textbf{observation space}, there are several levels of observability: there are easily observable states in a real factory that a sensor could essentially measure, for example temperature; there are also some states that are expensive and slow to observe, like some concentrations; there are also some hardly observable states, like the internal changes of some chemical reactions. In the experiments of this paper, we observe everything except the internal changes, forming into the Partially Observable Markov Decision Process (POMDP). 
To mimic how the actual process is controlled and run, we include 2 categories of actions in the \textbf{action space}: 1) manipulable variables. They are the setpoints in the typical control sense (e.g. the temperature of a cooling jacket outside a reactor). 2) input materials. Input materials could be the raw materials that go into the process like sugar, or the attributes of the materials like the concentration. Typically, they can be determined by the operator to optimize the process.
The environment \textbf{transitions} are modeled with differential equations with first principal or empirical approximations. With the help of domain experts, we managed to access and utilize the industrial data to validate our models and determine the corresponding parameters to align with real-world factories. There are several \textbf{reward functions} designed for each of the environments. In the experiments of this paper we only use the dense rewards that measure performance. More specifically, we want a successful control algorithm to be first and foremost safe, then more stable, efficient and productive. Since we have our \hyperref[sec:Baseline Algorithms]{baseline algorithms}, we compare the results of the reinforcement learning algorithms with our baselines. We define more efficient and more productive to be higher in average rewards. We define more stable to be less standard deviation in rewards (which only matters when the algorithms are efficient and productive enough, since a zero-yield function is stable on itself). We would like to clarify that "stable" or "stability" as used throughout this article does not refer to the notion of stable or stability in control theory.
Note that in real-life circumstances, efficiency means less product investment (e.g. less sugar and Biomass input as of BeerFMTEnv) and productivity means production yield (e.g. more beer produced as of BeerFMTEnv). But in this paper, our reward function takes both production investment and production yield into account, so we use reward as our single measurement criteria here.

In summary, the main contributions of this paper include:
(i) we build five novel manufacturing simulation environments with high fidelity to facilitate researches in these areas; (ii) we tune advanced control algorithms used in industry for the simulation environments, as comparable baselines; (iii) we benchmark popular online and offline, model-based and model-free reinforcement learning algorithms for future reference. In this work, we aim to build simulation environments, and encourage the community to utilize them, in order to find deep reinforcement learning algorithms that solve the real-life environments.

\section{Related Work}

Since the success of Deep Q-Learning (DQN) in Atari games \cite{DQN}, a variety of environments have been developed, including games~\cite{openaigym, dota2, wydmuch2018vizdoom, https://doi.org/10.48550/arxiv.2109.06780, LanctotEtAl2019OpenSpiel, cote18textworld}, kinematics~\cite{dm_control}, autonomous driving~\cite{airsim2017fsr}, recommendation systems~\cite{rohde2018recogym}, multi-agent collaborations~\cite{smac, gfootball}, networks~\cite{ns3gym}, and offline reinforcement learning data collections~\cite{d4rl}, which can be used to evaluate deep reinforcement learning algorithms.
However, there is a lack of established environments for process control in manufacturing. Considering the vast differences, it is difficult to determine whether a deep reinforcement learning algorithm, which works well in popular benchmarks such as Atari games and dm\_control, can be successfully adapted for a production setting owing to the domain discrepancy.

Atari games have discrete image observations and discrete actions, whereas SMPL has continuous observations and actions. Also in Atari games, the visual observations would not have a drastic change in one step. In chemical manufacturing, however, the pH (state) may not respond to a continuous flow of acid (action) after several minutes or even hours, but can also vary greatly with only a slight change in the concentration of such an input like in titration. As opposed to games where one can pause and resume, a factory cannot not wait for the computation to finish before taking the action. The runtime of control algorithms should also be taken into consideration. As compared to other Advanced Process Control (APC) algorithms, inference of deep reinforcement learning algorithms is faster. 

As compared to dm\_control (or other MuJoCo~\cite{mujoco} based physics simulation environments) which is also continuous in states and actions, our environments are still different. Firstly, each of the states in dm\_control has specific and accurate low-level physical semantics, like angle, coordinate, or velocity of a joint. It only 
records the transition of parts as a kinematical abstraction. Whereas in SMPL, the states are flow rates, the volume of liquid, the concentration of the solution, etc. Since "more is different"~\cite{moreisdifferent}, SMPL models things at a much larger scale, even though still built with differential equations. In SMPL, those differential equations are modeling physical, chemical and biological processes in manufacturing simulations, and many of them are empirical or phenomenological approximations, so distortions could be a big problem outside their range. To avoid distribution shift between simulations and actual factory transition dynamics, we need to restrict the action and state spaces to a relatively small range. 
Again, similar to the image represented Atari game states, the position and velocity of a dm\_control object would not change suddenly. SMPL, on the other hand, can sometimes change drastically.

To summarize, SMPL is challenging for learning algorithms that work well on Atari games or dm\_control simulations. Firstly, SMPL has qualitatively different dynamics. The rate of state change with respect to actions can vary widely. A small change in actions may result in a huge change in states, and the effect of a large change in actions may be delayed for hours. Secondly, as compared to games or robot walking, the punishment of failure is harsh in SMPL, due to security concerns similar to autonomous driving. Thirdly, we observed that the reinforcement learning algorithms tend to exploit the simulators. But due to the complexity of chemical and biomedical reactions, our simulations can easily break or occasionally reach an undefined state when the agent is exploring outside the reliable region. Therefore, compared to the \hyperref[tab:Offline and Online Experiment Results Part 1]{results} of reinforcement learning experiments, a fairly simple hand-tuned \hyperref[sec:PID]{Proportional–Integral–Derivative (PID) controller} can provide tolerable performance. With SMPL, we hope to enable deep reinforcement learning researchers to address this interesting gap.

There have also been efforts in applying reinforcement learning in manufacturing problems~\cite{he2021deep, 10.1145/3424311.3424326, govindaiah2021applying}. However, to the best of our knowledge, there has not been any other work that provides open-source standardized biochemical process control environments for reinforcement learning and advanced control. More typically, to utilize simulated environments for process control, researchers need to spend sufficient time to understand the underlying mathematical equations, prepare and further develop the environments. SMPL pitched those pain points for reinforcement learning researchers to develop their solutions for manufacturing environments.

\section{Environments}

The SMPL supplements several process control environments to the OpenAI Gym family \cite{openaigym}, which alleviates the pain of performing Deep Reinforcement Learning algorithms on them. Furthermore, we provided D4RL-like \cite{d4rl} wrappers for accompanied datasets, making offline reinforcement learning in those environments even smoother. Details of each of the environments can be found in Table~\ref{tab:Env and dataset Details}.

The transitions of the environments are based on Ordinary Differential Equations (ODEs) described in Appendices~\ref{sec:Environment Details}. The ODEs themselves model chemical, biological and mechanical transitions and reactions after taking an action from our action space, returning the state and evaluated reward. Originally from a control-theory perspective, we have adapted the transition dynamic to be time-invariant. In real life, the transition contains uncertainty due to the chemical, biological and mechanical process or the sensors' error, but as described in Section~\ref{sec:Limitations}, the noises are hard to model and cannot be simplified as Gaussian noises. The action space and state space are all continuous, with their safety constraints respectively.

\subsection{ReactorEnv}
This Continuous Stirred Tank Reactor (CSTR) process model is a representation of the most common container for a continuous reaction to take place. Even though we have already configured it for a particular reaction, it could easily be re-configured for other tasks (for example, change the cooling jacket to a heating jacket if the reaction is endothermic).

\begin{figure}[h!]
  \begin{center}
    \includegraphics[width=0.5\textwidth]{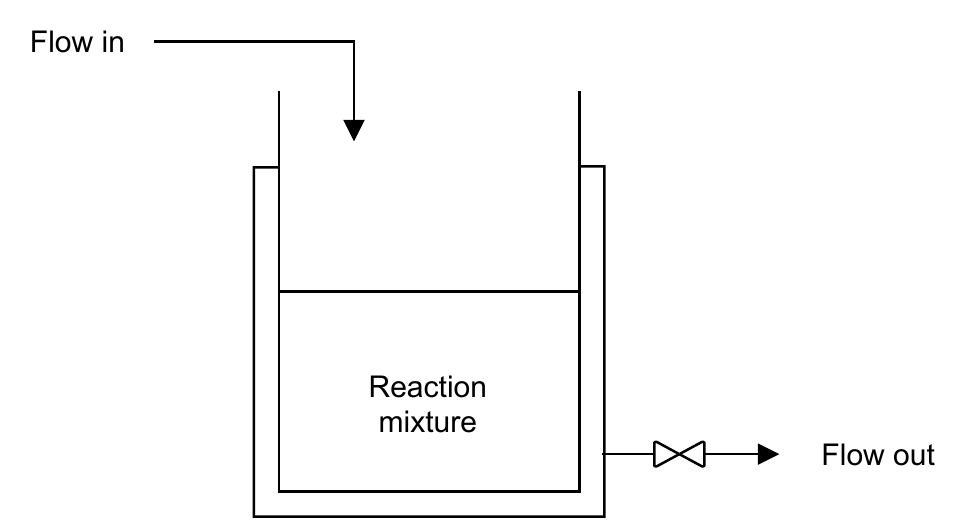}    
    \caption{Process flow diagram of the continuous manufacturing process} 
    \label{fig:process_flow_diagram}
  \end{center}
\end{figure}

A schematic diagram of a CSTR is presented in Figure~\ref{fig:process_flow_diagram}. In this article, we consider the case where a single first-order irreversible exothermic reaction of the form A → B takes place in the reactor. Because it is a continuous process, the reactants and the products are continuously fed and withdrawn from the reactor, respectively. Because the reaction is exothermic, thermal energy is removed from the reactor through a cooling jacket. The following assumptions are also taken in deriving the model:

\begin{itemize}
\item The reaction mixture is well mixed. This implies that there are no spatial variations in the reaction mixture.
\item Heat losses to the surroundings are negligible or nonexistent. 
\end{itemize}

The reaction details and the configurations can be found in Appendix~\ref{sec:appendices CSTR ReactorEnv}.


\subsection{AtropineEnv}
\begin{figure}[!ht]
  \begin{center}
    \includegraphics[width=0.9\textwidth]{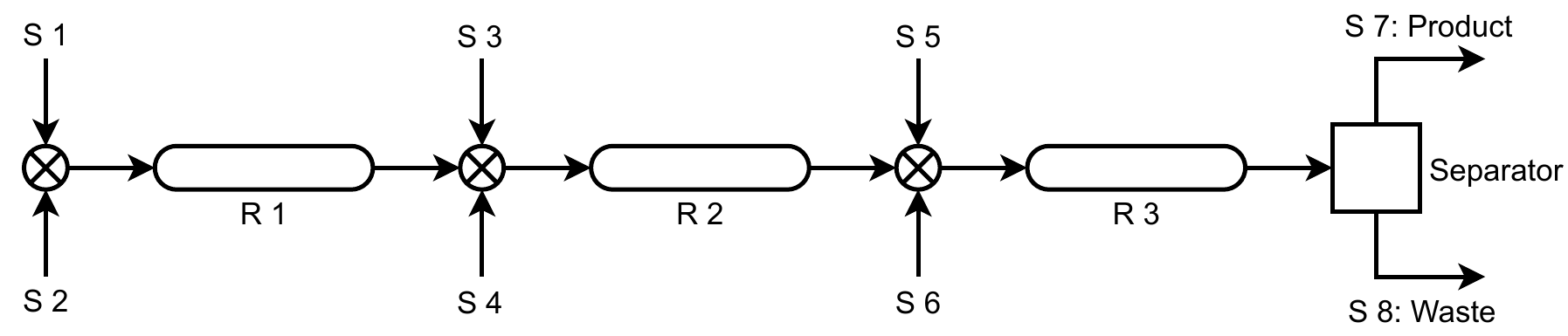}    
    \caption{Process flow diagram of the continuous manufacturing process. Details can be found in Table~\ref{tb:streams}} 
    \label{fig:atropine_flow}
  \end{center}
\end{figure}

Atropine is a common tropane alkaloid and anticholinergic medication used to treat certain types of nerve agents and pesticide poisonings as well as some types of slow heart rate and to decrease saliva production during surgery.  Figure~\ref{fig:atropine_flow} shows the process flow diagram for the atropine production process as presented in the work by Nikolakopoulou, von Andrian and Braatz \cite{nikolakopoulou2020}. This environment simulates a continuous-flow manufacturing process of atropine production . The simulation consists of three tubular reactors ($R_1$, $R_2$, $R_3$) in series and a liquid-liquid separator. Each reactor has a mixer proceeding it where the streams ($S_i$) containing the reactants are thoroughly mixed before being fed into the tubular reactor downstream of it. The end goal is to produce as much atropine as possible while keeping the reactor safe. Details can be found in Appendix~\ref{sec:appendices AtropineEnv}.

\subsection{mAbEnv}
This environment simulates a manufacturing process of monoclonal antibodies (mAbs), which are widely used for the treatment of autoimmune diseases, cancer, etc. According to a recent publication, mAbs also show promising results in the treatment of COVID-19 \cite{wang2020human}. Integrated continuous manufacturing of mAbs represents the state-of-the-art in mAb manufacturing and has attracted a lot of attention, because of the steady-state operations, high volumetric productivity, reduced equipment size and capital cost, etc. However, there is no existing mathematical model of the integrated manufacturing process and there is no optimal control algorithm for the entire integrated process. This project fills the knowledge gaps by first developing a mathematical model of the integrated continuous manufacturing process of mAbs.

The manufacturing process contains an upstream and a downstream process, and the end goal is to recover as much mAb as possible. Details can be found in Appendix~\ref{sec:appendices mAbEnv}.

\subsection{PenSimEnv}
\label{sec:PenSimEnv}
\begin{figure}[!ht]
  \begin{center}
    \includegraphics[width=0.8\textwidth]{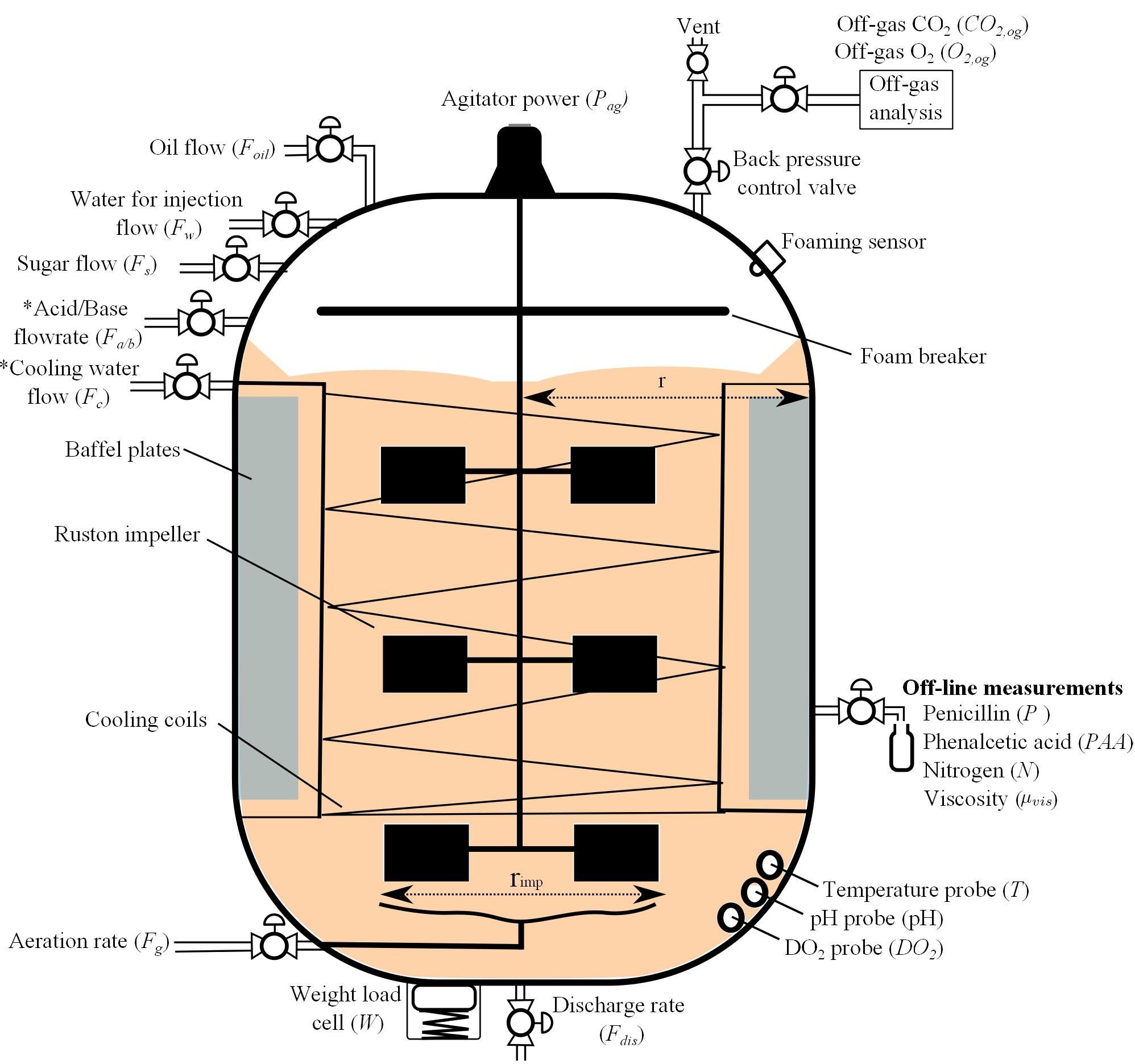}    
    \caption{A penicillin manufacturing process, from~\cite{goldrick2015development}} 
    \label{fig:penicillin manufacturing process}
  \end{center}
\end{figure}

Penicillin is the first-discovered group of antibiotics in human history. In this environment, we simulate the industrial-scale penicillium chrysogenum fermentation. The aim is to optimize the penicillin production per episode (or batch yield) while avoiding extreme inputs, outputs, or changes that can potentially break the reactor. Figure~\ref{fig:penicillin manufacturing process} shows the fermentation process. The simulation environment is based on this industrial-scale fed-batch fermentation simulation \cite{goldrick2015development}. Details can be found in Appendix~\ref{sec:appendices_PenSimEnv}. 


\subsection{BeerFMTEnv}
Although there are many different ways of beer production, typically, beer is produced in a fermentation unit where favorable conditions are created for the fermentation of the raw material. Optimal control of the beer fermentation process is a time-optimal control problem, which implies that the goal is to minimize the time required to completely ferment the raw materials. This environment provides a typical and simple enough simulation of the industry-level beer fermentation process. The manipulated variable for this environment is the reaction temperature. As mentioned earlier, the goal of this process is to reach the desired fermentation level within the shortest time possible. 

Since we just have a canonical industrial production formula that can only solve from a fixed initial state, we would only perform online reinforcement learning experiments. Details can be found in Appendix~\ref{sec:appendices_BeerFMTEnv}.

\subsection{Limitations}
\label{sec:Limitations}
Finally, while enabling machine learning practitioners and control engineers to explore more optimal control strategies and obtain actionable insights for real-world processes, the environments come with some limitations. Firstly, process variability and batch inconsistency is one of the major challenges in manufacturing, which cannot be simply described as gaussian noise. Adding more realistic noise and variability would make the simulations much closer to actual processes. Secondly, full accessibility to states may be difficult. For example, spectrophotometers are usually used for measuring chemical substances like concentrations, which could be costly. Last but not the least, state constraints are not fully considered. For example, due to safety and economic concerns, the next actions may not deviate too much from current actions. Although with the limitations described above, we believe that releasing the environments is still beneficial to researchers interested in applying reinforcement learning to control manufacturing processes. 

\section{Baseline Algorithms}
\label{sec:Baseline Algorithms}


In this section, we compare the performance of our deep reinforcement learning algorithms with a range of different baseline algorithms for each environment. Those baseline algorithms were tuned by experts on the environment. “We tuned the PID, MPC and EMPC by varying either the weights or the horizon, or both until we obtained closed-loop trajectories with little to no overshoot and fast response times. For offline reinforcement learning algorithms, we use the dataset generated by baseline algorithms as the training dataset. 

\textbf{Bayesian Optimization (BO) \cite{bayesian}}
\label{sec:Bayesian Optimization}

Bayesian optimization is a sample-efficient optimization method that focuses on solving the black-box optimization problem. In PenSimEnv, we maximize the penicillin yield by collecting 10 random trajectories as our start and then run 1000 Bayesian optimization searches. We use these 1000 trajectories as our offline reinforcement learning training dataset.

\textbf{Proportional–Integral–Derivative (PID) Control}
\label{sec:PID}
PID control is a classical feedback-based control loop that is widely used in industrial control systems. The controller itself is modulated, which computes and tries to minimize the difference between the current state and a set-point (target). 

\textbf{Advanced Process Control (APC)}
\label{sec:APC}
APC is a collection of control algorithms that is used in conjunction with PID to further improve the performance of industrial processes. Compared to reinforcement learning algorithms, APCs typically make use of a process model to predict the future evolution of process under the the selected control actions. In this article, we focus on two well known APC algorithms, namely setpoint tracking model predictive control (MPC) and economic model predictive control (EMPC).


\label{sec:MPC}
\textbf{MPC} also known as receding horizon control is an advanced process control algorithm that is able to handle systems with many states and constraints \cite{qin2003}. For this reason, it has had tremendous success in the chemical process industry. MPC requires a mathematical description of the process -- either from first principle or empirical -- to make predictions about the future evolution of the plant. In MPC, the economic performance objective of the process is translated to minimizing a quadratic cost function which measures the deviation of the system state and input from a desired setpoint. The setpoint is determined and updated by solving a steady-state economic optimization problem in a higher decision making layer known as real-time optimization (RTO). Thus, the economic performance of MPC is as good as the setpoint being tracked and the frequency of the setpoint update.  

\label{sec:EMPC}
\textbf{EMPC} is a variant of MPC that has gained tremendous attention in the process control community \cite{empc}. Compared to MPC, EMPC uses a more general cost function which generally reflects some economic indicator such as waste minimization or yield maximization. From a theoretical point of view, the performance of EMPC is no worse than that of MPC \cite{angeli2011}. This is because EMPC simply directly optimizes the process economics compared to MPC which requires that the economic objective be translated to a setpoint tracking objective.

There are still many challenges faced by MPC and its variants.

For example, when the dimension of the system model is too high, or when there are integer control inputs, it might be difficult to solve the MPC optimization problem in real time and within a reasonable time. EMPC faces a significantly higher computational challenges than MPC. This is because a more general dynamic economic optimization need to be solved in real time. Moreover, handling uncertainties in MPC or EMPC is a very challenging task \cite{mayne2016}. DRL on the other hand, works well under uncertainty with fast inference time.

\section{Experiments}
\label{sec:Experiments}

\begin{table}
\tablestyle[roman]
\centering

\scriptsize 
\begin{tabular}{c|rrrrrrrr}
\tbody
    \tsubheadstart
        \tsubhead \diagbox{Env}{Config} &
        \tsubhead Baseline &
        \tsubhead Traj \# &
        \tsubhead $\mu_r$ &
        \tsubhead $\sigma_r$ &
        \tsubhead a\_dim &  
        \tsubhead o\_dim &  
        \tsubhead max\_s &
        \tsubhead e\_r \\\hline
    ReactorEnv &MPC &125,000 &-3.7528 &60.4470 &2 &3 &100 &-1000 \\
    AtropineEnv &MPC &10,000 &-20.6094 &339.5105 &4 &39 &60 &-100000 \\
    PenSimEnv &BO &1000 &3.3071 &1.4673 &6 &9 &1150 &-100 \\
    mAbEnv &MPC &1000 &1322.5012 &174.7515 &9 &1970 &200 &-100 \\
    BeerFMTEnv & N/A & N/A & N/A & N/A &1 &8 &200 &-200 \\
\tend
\end{tabular}
\medskip

\caption{Env and Dataset Details. For each environment, from left to right: "Baseline" is the name of the baseline algorithm that the dataset is generated with; "Traj \#" is the number of trajectories; "$\mu_r$" is the mean of the rewards of the dataset; "$\sigma_r$ is the standard deviation of the rewards of the dataset; "a\_dim" (represents "action\_dim") is the dimension of action space in the environment; "o\_dim" (represents "observation\_dim") is the dimension of observation space in the environment; "max\_s" (represents "max\_steps") is the maximum number of steps possible in the environment; "e\_r" is called "error\_reward", which is given to a failed trajectory in the environment. Note that error\_reward always satisfies $error\_reward \leq r_{min} \times max\_steps$, where $r_{min}$ is the least possible reward for a step in the environment.}
\label{tab:Env and dataset Details}
\end{table}

\begin{table}
\tablestyle[roman]
\centering
\caption{Offline and Online Experiment Results Part 1}\label{tab:Offline and Online Experiment Results Part 1}
\scriptsize 

\begin{tabular}{c|rr|rr|rr|rr}
\theadstart
    \thead \diagbox{Algo}{Env} &
    \thead ReactorEnv & &
    \thead AtropineEnv & & 
    \thead PenSimEnv & &
    \thead mAbEnv & 
    \\
\tbody
    \tsubheadstart
        \tsubhead OfflineRL &
        \tsubhead $\mu_r$ &
        \tsubhead $\sigma_r$ &
        \tsubhead $\mu_r$ &
        \tsubhead $\sigma_r$ &
        \tsubhead $\mu_r$ &
        \tsubhead $\sigma_r$ &
        \tsubhead $\mu_r$ &
        \tsubhead $\sigma_r$ \\\hline
    \textit{\hyperref[sec:MPC]{MPC}} &\textit{-0.1912} &\textit{15.9328} &\textit{-20.6094} &\textit{339.5105} & N/A & N/A &\textit{1322.5012} &\textit{174.7515} \\
    \textit{\hyperref[sec:Bayesian Optimization]{GPEI}} & N/A & N/A & N/A & N/A &\textit{3.3071} &\textit{1.4673} & N/A & N/A \\
    \hyperref[sec:EMPC]{EMPC} & N/A & N/A & N/A & N/A & N/A & N/A &1314.1145 &221.4624 \\
    \hyperref[sec:PID]{PID} &-1.0823 &35.0924 & N/A & N/A & N/A & N/A & N/A & N/A \\
    PLAS~\cite{PLAS} &-0.1909 &27.9975 &\textbf{-4.8416} &6.7930 &2.0421 &4.7936 &1324.1982 &124.7369 \\
    PLASWithPert~\cite{PLAS} &-0.1638 &13.2464 &-17.5712 &6.1863 &\textbf{3.0056} &4.7353 &1357.0842 &122.1667 \\
    TD3~\cite{TD3} &-0.6428 &25.4631 &-51.3783 &\textbf{4.1409} &-1.2575 &11.1618 &708.1167 &571.8498 \\
    AWAC~\cite{AWAC} &\textbf{-0.0965} &\textbf{13.9237} &-8.4070 &10.8725 &2.3288 &4.6710 &\textbf{1374.2244} &\textbf{61.2931} \\
    BEAR~\cite{BEAR} &-13.4975 &111.0537 &-28.2412 &11.6400 &2.2086 &\textbf{4.4271} &1268.8020 &297.5198 \\
    BCQ~\cite{BCQ} &-2.0008 &43.3134 &-18.2633 &4.6889 &2.1341 &4.6710 &1247.0172 &313.2412 \\
    SAC~\cite{SAC} &-0.7120 &18.5272 &-181.0055 &5.4775 &2.5961 &4.8317 &1197.1507 &503.1202 \\
    DDPG~\cite{DDPG} &-1.3219 &62.6639 &-203.0525 &5.7211 &-1.9732 &13.9105 &711.1741 &574.0687 \\
    CQL~\cite{CQL} &-2.5266 &79.5985 &-13.9241 &8.9072 &2.5543 &4.5791 &1254.5728 &233.4812 \\
    COMBO~\cite{COMBO} &-7.9475 &84.2047 &-30.1022 &11.2495 &3.1368 &4.5718 &1165.3812 &440.0366 \\
    MOPO~\cite{MOPO} &-8.8802 &94.0311 &-90.8561 &52.7119 &1.4345 &5.0138 &-100.0000 &0.0000 \\
    BC~\cite{BC} &-3.5765 &59.8082 &-5.4473 &44.1707 &0.7440 &4.5213 &1322.8184 &111.1911 \\
    \hline
    \tsubheadstart
        \tsubhead OnlineRL & & & & & & & & \\\hline
    PPO~\cite{PPO} &-31.9667 &180.9413 &-67.4412 &2595.8698 &2.5231 &4.6745 &-92.7862 &137.6964 \\
    A3C~\cite{A2CA3C} &-1000.0000 &0.0000 &-50.2008 &2239.2415 &-0.8551 &9.3091 &-88.8452 &272.8659 \\
    ARS~\cite{ARS} &-1000.0000 &0.0000 &-83333.3879 &37267.6776 & & & & \\
    IMPALA~\cite{IMPALA} &-1000.0000 &0.0000 &-100000.0000 &0.0000 &-2.0575 &14.1869 &-94.3842 &184.2320 \\
    PG~\cite{PG} &-140.8971 &311.5265 &-102.5358 &3194.3687 &2.1039 &4.3759 &-86.1046 &241.7652 \\
    SAC &-113.3525 &293.3941 &-16.4201 &1280.6835 &-2.3044 &15.0070 &-90.1491 &233.4410 \\
    DDPG &-97.6231 &279.8284 &-141.0357 &302.8068 &-0.8373 &9.1140 &-95.1542 &288.9491 \\
\tend

\end{tabular}

\bigskip

\begin{tabular}{c|rr}
\tbody
    \tsubheadstart
        \tsubhead \diagbox{OnlineRL}{BeerFMTEnv} &
        \tsubhead $\mu_r$ &
        \tsubhead $\sigma_r$ \\\hline
    PPO &1.0688 &20.7031 \\
    A3C &\textbf{0.8706} &20.3378 \\
    ARS &-200.0000 &0.0000 \\
    IMPALA &-1.0371 &14.5516 \\
    PG &-1.9900 &\textbf{14.0014} \\
    SAC &-1.9753 &14.0415 \\
    DDPG &-2.1842 &15.3056 \\
\tend
\end{tabular}

\end{table}

In this section, we demonstrate the results of our experiments. In the experiments, each offline reinforcement learning algorithm is trained for 500 \hyperref[sec:OfflineRL Data Collection]{epochs}, and each online reinforcement learning algorithm is trained for 2 million environment steps. These experiments can be treated as a benchmarking baseline for future research in achieving higher stability, efficiency and productivity. 
For the environments, we mimic the design of the OpenAI Gym so that we can easily train and test a variety of reinforcement learning algorithms on them. The associated code can be accessed with the provided code.

\subsection{Offline Reinforcement Learning}
\label{sec:Offline Reinforcement Learning}
The simulations are meant to mimic and abstract existing factories, and the traditional manufacturing plants are controlled by human workers or threshold functions. Luckily, most of the states and actions are recorded by sensors and workers. A natural thought would be to use those historical data to learn better control algorithms, and a successful control algorithm should be more stable, efficient and productive. Offline reinforcement learning is a possible approach since it can purely learn from historical data. As in real life, it is infeasible to train RL agents on actual manufacturing production lines due to safety and economic concerns. With offline reinforcement learning, a control policy can be developed by learning from offline data collected from manual control or APC.

\phantomsection
\label{sec:OfflineRL Data Collection}
Table~\ref{tab:Offline and Online Experiment Results Part 1} shows the experiment results. Each of the four simulation environments has an expert control algorithm which is manually tuned by experts. This expert control algorithm can provide successful controls in non-extreme cases, with potentially low efficiency due to online optimization and computation complexity.

We set a maximum time limit for each of the environments. If a trajectory generated by an algorithm in an environment ends before the maximum time limit due to an error reported by the environment, or if an action/observation goes beyond the allowed limit, then this trajectory is marked as failed; otherwise, this trajectory is marked as succeeded. A failed trajectory receives a negatively large reward as the punishment (the name of which is error\_reward). Since the initial states are not fixed, a successful algorithm needs to adapt to a wide range of situations. 

Specifically as Table~\ref{tab:Env and dataset Details} shows, ReactorEnv, AtropineEnv and mAbEnv all have MPC as their baseline algorithm. PenSimEnv uses a version of Gaussian Process based Bayesian Optimization (GPEI) developed in-house. We generate datasets with these algorithms, to train our reinforcement learning algorithms offline. Furthermore, ReactorEnv has a PID controller and mAbEnv has an EMPC controller for comparison. We did not perform offline reinforcement learning on the BeerFMTEnv since it only has a static rule as the baseline. The amount of data sampled differently because of the cost of sampling. The environments are all drastically different from each other and cover different types of manufacturing processes.

D4RL \cite{d4rl} is one of the most widely used benchmarks in the offline reinforcement learning community. Thus, we make the generation of the dataset in both D4RL and Torch format with any control algorithm possible in our library, and only a few lines of code would be sufficient.

Thanks to D3RLPY \cite{d3rlpy}, we can perform batched parallel training with a little engineering. The results are shown in Table~\ref{tab:Offline and Online Experiment Results Part 1}:

On ReactorEnv, only Advantage Weighted Actor-Critic (AWAC) is able to outperform the MPC baseline. In terms of average rewards, Policy in the Latent Action Space (PLAS) and PLAS with perturbation slightly go beyond our baseline.

On AtropineEnv, PLAS has the maximum average rewards, while Twin Delayed Deep Deterministic Policy Gradients (TD3) has the smallest standard deviation of rewards. Moreover, AWAC, Batch-Constrained Q-learning (BCQ), Conservative Q-Learning (CQL) and Behavior Cloning (BC) are able to provide performances beyond our Model Predictive Control (MPC) baseline. Among them, we could say that PLAS, BC and AWAC solved the AtropineEnv better than the MPC baseline for they have a 50\% more average and a 50\% less standard deviation of rewards.

On PenSimEnv, it is also PLAS that has the highest average rewards, while Bootstrapping Error Accumulation Reduction (BEAR) has the lowest standard deviation of rewards. However, none of the offline reinforcement learning algorithms can produce average rewards higher than the Gaussian Process based Bayesian Optimization (GPEI) baseline.

On mAbEnv, AWAC wins the highest maximum average and the smallest standard deviation of rewards. PLAS with perturbation, PLAS and BC all have slightly better performances compared to our baseline MPC in terms of average and standard deviation. 

PLAS \cite{PLAS} seems to achieve a better performance than the baseline in many environments. Moreover, on mAbEnv, PLAS with perturbation is better, because we only have 1000 MPC trajectories. For AtropineEnv, PLAS without perturbation has higher performance, since AtropineEnv has 10,000 trajectories. On ReactorEnv, PLAS with our without perturbation has a very similar performance, and there are 125,000 trajectories. From these experiments, we observe that the perturbation layer, which aims to sample generalized action out of the training dataset, could harm the performance when the training dataset is too small, but can improve the performance when the dataset is large enough.


AWAC, with its ability to utilize sub-optimal data-points based on traditional advantage actor-critic algorithms~\cite{A2CA3C}, is able to thrive in both data-abundant (ReactorEnv) and data-scarce (mAbEnv) scenarios. The fact that it outperforms all other offline reinforcement learning algorithms makes its data-utilization trick worthwhile for further exploration.

Deep Deterministic Policy Gradients (DDPG) has poor performance in most environments, as compared to other q-learning variants, even when provided dense rewards~\cite{problemwithddpg}. Some researches shows that without careful tuning of the hyperparameters, DDPG and Soft Actor-Critic (SAC) tend to largely overestimate the q-value during the training, and the exploding imitation value estimation leads to an inescapable extrapolation error \cite{BCQ}. However, when exploration is allowed, DDPG could have a \hyperref[sec:Online DDPG]{slightly better performance}. Compared to DDPG and SAC, which were not designed for offline reinforcement learning, BCQ claims to be able to improve on restricting the action spaces, but it still suffers from the q-value overestimation in PenSimEnv and mAbEnv.

There are some algorithms (like A3C, ARS, IMPALA in ReactorEnv and IMPALA in AtropineEnv) that always tend to break the simulation by either going out of the limit or reaching an unacceptable state. With careful tuning, their performance could be improved.

\subsection{Online Reinforcement Learning}
\label{sec:Online Reinforcement Learning}
In real-life use cases, we cannot directly apply q-learning or policy gradient on the producing plant for a hundred thousand episodes to train a good RL agent. However, we have the simulations anyways, we would like to perform online reinforcement learning experiments on those simulations, and the results can serve as baselines.

We utilize the Ray \cite{ray} library, with its RLlib \cite{rllib} and Tune \cite{tune} components to parallelize the training. With a little tuning and a limited budget of training time, the performance of online reinforcement learning algorithms is all lower than the offline reinforcement learning algorithms. We believe that the exploration in SMPL is hard in general, and the guidance provided by expert algorithms can provide successful controls in most cases (with potentially low efficiency).

\label{sec:Online DDPG}
Note that compared to the offline reinforcement learning experiments, SAC shows worse results. DDPG, on the other hand, is able to show a slightly better result in PenSimEnv. The hypothesis could be that, due to the very slight change in rewards (the dataset of PenSimEnv has a very small standard deviation of rewards, only 1.4673), the over-estimation problem of DDPG might be fixed by exhaustive exploration.

\section{Conclusions and Future Work}


In our work, we introduced five simulation environments that covered a wide range of manufacturing processes. Corresponding with the environments, we provided expert-tuned control algorithms that are employed in factories. Based on the built environments and baselines, we tried offline and online, model-based and model-free reinforcement learning algorithms.

From the experiment results, we suggest utilizing offline reinforcement learning algorithms and learning from the baseline-generated samples as a starting point. 


Only on AtropineEnv, a few reinforcement learning algorithms show a significant improvement as compared to baselines, while on other environments the reinforcement learning approaches are only marginally better, if at all.
Therefore, targeted research, specifically designed or carefully tuned algorithms might be a prerequisite to succeed on our environments, for 1. our environments are challenging and complex (especially mAbEnv, PenSimEnv and ReactorEnv) 2. there are significant differences between our environments and the environments where the reinforcement learning algorithms are originally designed or often experimented. 

We will continue to work on the existing environments to find a more stable, efficient and productive algorithm. Experiments already show us that directly applying existing algorithms might not be a solution, so further research on how to sample safe actions from empirical results given only historical data might be a good starting point. The overall good performance of PLAS hints that digging into shaping the latent action space might be a good idea.

Another one of our goals is to find an algorithm that can perform well in all the environments with little modification, which can be useful if a plant decides to change a reactor tank, or we want to adopt a trained algorithm for a new plant. We would like to develop existing meta-learning algorithms like \cite{nam2022skillbased, luo2022adapt}. 

In addition to the existing configurations, we would also like to add more tunable parameters to the environments that can represent different types of manufacturing processes. More simulation environments, on top of the existing five, would be added to the family once finished. 

We would actively develop and maintain this library, to better serve the reinforcement learning community.


\bibliographystyle{ieeetr}
\bibliography{ref}

\begin{thebibliography}{10}

\bibitem{thomas2020annual}
D.~S. Thomas {\em et~al.}, ``Annual report on us manufacturing industry
  statistics: 2020,'' 2020.

\bibitem{mahadevan1998optimizing}
S.~Mahadevan and G.~Theocharous, ``Optimizing production manufacturing using
  reinforcement learning.,'' in {\em FLAIRS conference}, vol.~372, p.~377,
  1998.

\bibitem{OVERBECK2021170}
L.~Overbeck, A.~Hugues, M.~C. May, A.~Kuhnle, and G.~Lanza, ``Reinforcement
  learning based production control of semi-automated manufacturing systems,''
  {\em Procedia CIRP}, vol.~103, pp.~170--175, 2021.
\newblock 9th CIRP Global Web Conference – Sustainable, resilient, and agile
  manufacturing and service operations : Lessons from COVID-19.

\bibitem{DQN}
V.~Mnih, K.~Kavukcuoglu, D.~Silver, A.~A. Rusu, J.~Veness, M.~G. Bellemare,
  A.~Graves, M.~Riedmiller, A.~K. Fidjeland, G.~Ostrovski, {\em et~al.},
  ``Human-level control through deep reinforcement learning,'' {\em nature},
  vol.~518, no.~7540, pp.~529--533, 2015.

\bibitem{openaigym}
G.~Brockman, V.~Cheung, L.~Pettersson, J.~Schneider, J.~Schulman, J.~Tang, and
  W.~Zaremba, ``Openai gym,'' {\em arXiv preprint arXiv:1606.0154}, 2016.

\bibitem{dota2}
{OpenAI}, {:}, C.~Berner, G.~Brockman, B.~Chan, V.~Cheung, P.~Dębiak,
  C.~Dennison, D.~Farhi, Q.~Fischer, S.~Hashme, C.~Hesse, R.~Józefowicz,
  S.~Gray, C.~Olsson, J.~Pachocki, M.~Petrov, H.~P. d.~O. Pinto, J.~Raiman,
  T.~Salimans, J.~Schlatter, J.~Schneider, S.~Sidor, I.~Sutskever, J.~Tang,
  F.~Wolski, and S.~Zhang, ``Dota 2 with large scale deep reinforcement
  learning,'' {\em arXiv preprint arXiv:1912.06680}, 2019.

\bibitem{wydmuch2018vizdoom}
M.~Wydmuch, M.~Kempka, and W.~Ja{\'s}kowski, ``Vizdoom competitions: Playing
  doom from pixels,'' {\em IEEE Transactions on Games}, 2018.

\bibitem{https://doi.org/10.48550/arxiv.2109.06780}
D.~Hafner, ``Benchmarking the spectrum of agent capabilities,'' {\em arXiv
  preprint arXiv:2109.06780}, 2021.

\bibitem{LanctotEtAl2019OpenSpiel}
M.~Lanctot, E.~Lockhart, J.-B. Lespiau, V.~Zambaldi, S.~Upadhyay,
  J.~P\'{e}rolat, S.~Srinivasan, F.~Timbers, K.~Tuyls, S.~Omidshafiei,
  D.~Hennes, D.~Morrill, P.~Muller, T.~Ewalds, R.~Faulkner, J.~Kram\'{a}r,
  B.~D. Vylder, B.~Saeta, J.~Bradbury, D.~Ding, S.~Borgeaud, M.~Lai,
  J.~Schrittwieser, T.~Anthony, E.~Hughes, I.~Danihelka, and J.~Ryan-Davis,
  ``{OpenSpiel}: A framework for reinforcement learning in games,'' {\em CoRR},
  vol.~abs/1908.09453, 2019.

\bibitem{cote18textworld}
M.-A. C\^ot\'e, A.~K\'ad\'ar, X.~Yuan, B.~Kybartas, T.~Barnes, E.~Fine,
  J.~Moore, R.~Y. Tao, M.~Hausknecht, L.~E. Asri, M.~Adada, W.~Tay, and
  A.~Trischler, ``Textworld: A learning environment for text-based games,''
  {\em CoRR}, vol.~abs/1806.11532, 2018.

\bibitem{dm_control}
S.~Tunyasuvunakool, A.~Muldal, Y.~Doron, S.~Liu, S.~Bohez, J.~Merel, T.~Erez,
  T.~Lillicrap, N.~Heess, and Y.~Tassa, ``dm{\_}control: Software and tasks for
  continuous control,'' {\em Software Impacts}, vol.~6, p.~100022, nov 2020.

\bibitem{airsim2017fsr}
S.~Shah, D.~Dey, C.~Lovett, and A.~Kapoor, ``Airsim: High-fidelity visual and
  physical simulation for autonomous vehicles,'' in {\em Field and Service
  Robotics}, 2017.

\bibitem{rohde2018recogym}
D.~Rohde, S.~Bonner, T.~Dunlop, F.~Vasile, and A.~Karatzoglou, ``Recogym: A
  reinforcement learning environment for the problem of product recommendation
  in online advertising,'' {\em arXiv preprint arXiv:1808.00720}, 2018.

\bibitem{smac}
M.~Samvelyan, T.~Rashid, C.~S. de~Witt, G.~Farquhar, N.~Nardelli, T.~G.~J.
  Rudner, C.-M. Hung, P.~H.~S. Torr, J.~Foerster, and S.~Whiteson, ``{The}
  {StarCraft} {Multi}-{Agent} {Challenge},'' {\em CoRR}, vol.~abs/1902.04043,
  2019.

\bibitem{gfootball}
K.~Kurach, A.~Raichuk, P.~Sta{\'n}czyk, M.~Zaj{\k{a}}c, O.~Bachem, L.~Espeholt,
  C.~Riquelme, D.~Vincent, M.~Michalski, O.~Bousquet, {\em et~al.}, ``Google
  research football: A novel reinforcement learning environment,'' in {\em
  Proceedings of the AAAI Conference on Artificial Intelligence}, vol.~34,
  pp.~4501--4510, 2020.

\bibitem{ns3gym}
P.~Gaw{\l}owicz and A.~Zubow, ``{ns-3 meets OpenAI Gym: The Playground for
  Machine Learning in Networking Research},'' in {\em {ACM International
  Conference on Modeling, Analysis and Simulation of Wireless and Mobile
  Systems (MSWiM)}}, November 2019.

\bibitem{d4rl}
J.~Fu, A.~Kumar, O.~Nachum, G.~Tucker, and S.~Levine, ``D4rl: Datasets for deep
  data-driven reinforcement learning,'' {\em arXiv preprint arXiv:2004.07219},
  2020.

\bibitem{mujoco}
E.~Todorov, T.~Erez, and Y.~Tassa, ``Mujoco: A physics engine for model-based
  control,'' in {\em 2012 IEEE/RSJ International Conference on Intelligent
  Robots and Systems}, pp.~5026--5033, IEEE, 2012.

\bibitem{moreisdifferent}
P.~W. Anderson, ``More is different,'' {\em Science}, vol.~177, no.~4047,
  pp.~393--396, 1972.

\bibitem{he2021deep}
Z.~He, K.-P. Tran, S.~Thomassey, X.~Zeng, J.~Xu, and C.~Yi, ``A deep
  reinforcement learning based multi-criteria decision support system for
  optimizing textile chemical process,'' {\em Computers in Industry}, vol.~125,
  p.~103373, 2021.

\bibitem{10.1145/3424311.3424326}
L.~Wang and Y.~Wang, ``Application of machine learning for process control in
  semiconductor manufacturing,'' in {\em Proceedings of the 2020 International
  Conference on Internet Computing for Science and Engineering}, ICICSE '20,
  (New York, NY, USA), p.~109–111, Association for Computing Machinery, 2020.

\bibitem{govindaiah2021applying}
S.~Govindaiah and M.~D. Petty, ``Applying reinforcement learning to plan
  manufacturing material handling,'' {\em Discover Artificial Intelligence},
  vol.~1, no.~1, pp.~1--33, 2021.

\bibitem{nikolakopoulou2020}
A.~Nikolakopoulou, M.~von Andrian, and R.~D. Braatz, ``Fast model predictive
  control of startup of a compact modular reconfigurable system for
  continuous-flow pharmaceutical manufacturing,'' in {\em 2020 American Control
  Conference (ACC)}, pp.~2778--2783, IEEE, 2020.

\bibitem{wang2020human}
C.~Wang, W.~Li, D.~Drabek, N.~M. Okba, R.~van Haperen, A.~D. Osterhaus, F.~J.
  van Kuppeveld, B.~L. Haagmans, F.~Grosveld, and B.-J. Bosch, ``A human
  monoclonal antibody blocking sars-cov-2 infection,'' {\em Nature
  communications}, vol.~11, no.~1, pp.~1--6, 2020.

\bibitem{goldrick2015development}
S.~Goldrick, A.~{\c{S}}tefan, D.~Lovett, G.~Montague, and B.~Lennox, ``The
  development of an industrial-scale fed-batch fermentation simulation,'' {\em
  Journal of biotechnology}, vol.~193, pp.~70--82, 2015.

\bibitem{bayesian}
J.~Mo{\v{c}}kus, ``On bayesian methods for seeking the extremum,'' in {\em
  Optimization techniques IFIP technical conference}, pp.~400--404, Springer,
  1975.

\bibitem{qin2003}
S.~J. Qin and T.~A. Badgwell, ``A survey of industrial model predictive control
  technology,'' {\em Control engineering practice}, vol.~11, no.~7,
  pp.~733--764, 2003.

\bibitem{empc}
J.~B. Rawlings, D.~Angeli, and C.~N. Bates, ``Fundamentals of economic model
  predictive control,'' in {\em 2012 IEEE 51st IEEE Conference on Decision and
  Control (CDC)}, pp.~3851--3861, 2012.

\bibitem{angeli2011}
D.~Angeli, R.~Amrit, and J.~B. Rawlings, ``On average performance and stability
  of economic model predictive control,'' {\em IEEE transactions on automatic
  control}, vol.~57, no.~7, pp.~1615--1626, 2011.

\bibitem{mayne2016}
D.~Mayne, ``Robust and stochastic model predictive control: Are we going in the
  right direction?,'' {\em Annual Reviews in Control}, vol.~41, pp.~184--192,
  2016.

\bibitem{PLAS}
W.~Zhou, S.~Bajracharya, and D.~Held, ``Plas: Latent action space for offline
  reinforcement learning,'' {\em arXiv preprint arXiv:2011.07213}, 2020.

\bibitem{TD3}
S.~Fujimoto, H.~Hoof, and D.~Meger, ``Addressing function approximation error
  in actor-critic methods,'' in {\em International conference on machine
  learning}, pp.~1587--1596, PMLR, 2018.

\bibitem{AWAC}
A.~Nair, A.~Gupta, M.~Dalal, and S.~Levine, ``Awac: Accelerating online
  reinforcement learning with offline datasets,'' {\em arXiv preprint
  arXiv:2006.09359}, 2020.

\bibitem{BEAR}
A.~Kumar, J.~Fu, M.~Soh, G.~Tucker, and S.~Levine, ``Stabilizing off-policy
  q-learning via bootstrapping error reduction,'' {\em Advances in Neural
  Information Processing Systems}, vol.~32, 2019.

\bibitem{BCQ}
S.~Fujimoto, D.~Meger, and D.~Precup, ``Off-policy deep reinforcement learning
  without exploration,'' in {\em International conference on machine learning},
  pp.~2052--2062, PMLR, 2019.

\bibitem{SAC}
T.~Haarnoja, A.~Zhou, K.~Hartikainen, G.~Tucker, S.~Ha, J.~Tan, V.~Kumar,
  H.~Zhu, A.~Gupta, P.~Abbeel, {\em et~al.}, ``Soft actor-critic algorithms and
  applications,'' {\em arXiv preprint arXiv:1812.05905}, 2018.

\bibitem{DDPG}
T.~P. Lillicrap, J.~J. Hunt, A.~Pritzel, N.~Heess, T.~Erez, Y.~Tassa,
  D.~Silver, and D.~Wierstra, ``Continuous control with deep reinforcement
  learning,'' {\em arXiv preprint arXiv:1509.02971}, 2015.

\bibitem{CQL}
A.~Kumar, A.~Zhou, G.~Tucker, and S.~Levine, ``Conservative q-learning for
  offline reinforcement learning,'' {\em Advances in Neural Information
  Processing Systems}, vol.~33, pp.~1179--1191, 2020.

\bibitem{COMBO}
T.~Yu, A.~Kumar, R.~Rafailov, A.~Rajeswaran, S.~Levine, and C.~Finn, ``Combo:
  Conservative offline model-based policy optimization,'' {\em Advances in
  neural information processing systems}, vol.~34, pp.~28954--28967, 2021.

\bibitem{MOPO}
T.~Yu, G.~Thomas, L.~Yu, S.~Ermon, J.~Y. Zou, S.~Levine, C.~Finn, and T.~Ma,
  ``Mopo: Model-based offline policy optimization,'' {\em Advances in Neural
  Information Processing Systems}, vol.~33, pp.~14129--14142, 2020.

\bibitem{BC}
D.~A. Pomerleau, ``Alvinn: An autonomous land vehicle in a neural network,''
  {\em Advances in neural information processing systems}, vol.~1, 1988.

\bibitem{PPO}
J.~Schulman, F.~Wolski, P.~Dhariwal, A.~Radford, and O.~Klimov, ``Proximal
  policy optimization algorithms,'' {\em arXiv preprint arXiv:1707.06347},
  2017.

\bibitem{A2CA3C}
V.~Mnih, A.~P. Badia, M.~Mirza, A.~Graves, T.~Lillicrap, T.~Harley, D.~Silver,
  and K.~Kavukcuoglu, ``Asynchronous methods for deep reinforcement learning,''
  in {\em International conference on machine learning}, pp.~1928--1937, PMLR,
  2016.

\bibitem{ARS}
H.~Mania, A.~Guy, and B.~Recht, ``Simple random search provides a competitive
  approach to reinforcement learning,'' {\em arXiv preprint arXiv:1803.07055},
  2018.

\bibitem{IMPALA}
L.~Espeholt, H.~Soyer, R.~Munos, K.~Simonyan, V.~Mnih, T.~Ward, Y.~Doron,
  V.~Firoiu, T.~Harley, I.~Dunning, {\em et~al.}, ``Impala: Scalable
  distributed deep-rl with importance weighted actor-learner architectures,''
  in {\em International conference on machine learning}, pp.~1407--1416, PMLR,
  2018.

\bibitem{PG}
R.~S. Sutton, D.~McAllester, S.~Singh, and Y.~Mansour, ``Policy gradient
  methods for reinforcement learning with function approximation,'' in {\em
  Proceedings of the 12th International Conference on Neural Information
  Processing Systems}, NIPS'99, (Cambridge, MA, USA), p.~1057–1063, MIT
  Press, 1999.

\bibitem{d3rlpy}
M.~I. Takuma~Seno, ``d3rlpy: An offline deep reinforcement library,'' in {\em
  NeurIPS 2021 Offline Reinforcement Learning Workshop}, December 2021.

\bibitem{problemwithddpg}
G.~Matheron, O.~Sigaud, and N.~Perrin, ``The problem with {\{}ddpg{\}}:
  understanding failures in deterministic environments with sparse rewards,''
  {\em arXiv preprint arXiv:1911.11679}, 2020.

\bibitem{ray}
P.~Moritz, R.~Nishihara, S.~Wang, A.~Tumanov, R.~Liaw, E.~Liang, M.~Elibol,
  Z.~Yang, W.~Paul, M.~I. Jordan, {\em et~al.}, ``Ray: A distributed framework
  for emerging $\{$AI$\}$ applications,'' in {\em 13th USENIX Symposium on
  Operating Systems Design and Implementation (OSDI 18)}, pp.~561--577, 2018.

\bibitem{rllib}
E.~Liang, R.~Liaw, R.~Nishihara, P.~Moritz, R.~Fox, K.~Goldberg, J.~Gonzalez,
  M.~Jordan, and I.~Stoica, ``Rllib: Abstractions for distributed reinforcement
  learning,'' in {\em International Conference on Machine Learning},
  pp.~3053--3062, PMLR, 2018.

\bibitem{tune}
R.~Liaw, E.~Liang, R.~Nishihara, P.~Moritz, J.~E. Gonzalez, and I.~Stoica,
  ``Tune: A research platform for distributed model selection and training,''
  {\em arXiv preprint arXiv:1807.05118}, 2018.

\bibitem{nam2022skillbased}
T.~Nam, S.-H. Sun, K.~Pertsch, S.~J. Hwang, and J.~J. Lim, ``Skill-based
  meta-reinforcement learning,'' {\em arXiv preprint arXiv:2204.11828}, 2022.

\bibitem{luo2022adapt}
F.-M. Luo, S.~Jiang, Y.~Yu, Z.~Zhang, and Y.-F. Zhang, ``Adapt to environment
  sudden changes by learning a context sensitive policy,'' in {\em Proceedings
  of the AAAI Conference on Artificial Intelligence, Virtual Event}, 2022.

\bibitem{schiesser2012}
W.~E. Schiesser, {\em The numerical method of lines: integration of partial
  differential equations}.
\newblock Elsevier, 2012.

\bibitem{KONTORAVDI2008}
C.~Kontoravdi, S.~P. Asprey, E.~N. Pistikopoulos, and A.~Mantalaris,
  ``Application of global sensitivity analysis to determine goals for design of
  experiments: An example study on antibody-producing cell cultures,'' {\em
  Biotechnology Progress}, vol.~21, p.~1128–1135, Sep 2008.

\bibitem{KONTORAVDI2010}
C.~Kontoravdi, E.~N. Pistikopoulos, and A.~Mantalaris, ``Systematic development
  of predictive mathematical models for animal cell cultures,'' {\em Computers
  \& Chemical Engineering}, vol.~34, p.~1192–1198, Aug 2010.

\bibitem{gomis2020model}
J.~Gomis-Fons, H.~Schwarz, L.~Zhang, N.~Andersson, B.~Nilsson, A.~Castan,
  A.~Solbrand, J.~Stevenson, and V.~Chotteau, ``Model-based design and control
  of a small-scale integrated continuous end-to-end mab platform,'' {\em
  Biotechnology progress}, vol.~36, no.~4, p.~e2995, 2020.

\bibitem{PAPATHANASIOU2017}
M.~M. Papathanasiou, A.~L. Quiroga-Campano, F.~Steinebach, M.~Elviro,
  A.~Mantalaris, and E.~N. Pistikopoulos, ``Advanced model-based control
  strategies for the intensification of upstream and downstream processing in
  mab production,'' {\em Biotechnology Progress}, vol.~33, p.~966–988, Apr
  2017.

\bibitem{VILLIGER2016}
T.~K. Villiger, E.~Scibona, M.~Stettler, H.~Broly, M.~Morbidelli, and M.~Soos,
  ``Controlling the time evolution of mab n-linked glycosylation - part ii:
  Model-based predictions,'' {\em Biotechnology Progress}, vol.~32,
  p.~1135–1148, Jul 2016.

\bibitem{jimenez2016}
I.~Jimenez~del Val, Y.~Fan, and D.~Weilguny, ``Dynamics of immature mab
  glycoform secretion during cho cell culture: An integrated modelling
  framework,'' {\em Biotechnology Journal}, vol.~11, p.~610–623, Feb 2016.

\bibitem{clincke2013}
M.~Clincke, C.~Mölleryd, P.~K. Samani, E.~Lindskog, E.~Fäldt, K.~Walsh, and
  V.~Chotteau, ``Very high density of chinese hamster ovary cells in perfusion
  by alternating tangential flow or tangential flow filtration in wave
  bioreactor™—part ii: Applications for antibody production and
  cryopreservation,'' {\em Biotechnology Progress}, vol.~29, p.~768–777, May
  2013.

\bibitem{dizaji2016minor}
N.~F. Dizaji, {\em Minor Whey Protein Purification Using Ion-Exchange Column
  Chromatography}.
\newblock PhD thesis, The University of Western Ontario, 2016.

\bibitem{perez2009igg}
E.~X. Perez-Almodovar and G.~Carta, ``Igg adsorption on a new protein a
  adsorbent based on macroporous hydrophilic polymers. i. adsorption
  equilibrium and kinetics,'' {\em Journal of Chromatography A}, vol.~1216,
  no.~47, pp.~8339--8347, 2009.

\bibitem{andersson2019}
J.~A. Andersson, J.~Gillis, G.~Horn, J.~B. Rawlings, and M.~Diehl, ``Casadi: a
  software framework for nonlinear optimization and optimal control,'' {\em
  Mathematical Programming Computation}, vol.~11, no.~1, pp.~1--36, 2019.

\bibitem{antibody2020antibody}
A.~Society, ``Antibody therapeutics approved or in regulatory review in the eu
  or us,'' 2020.

\bibitem{kaplon2020antibodies}
H.~Kaplon, M.~Muralidharan, Z.~Schneider, and J.~M. Reichert, ``Antibodies to
  watch in 2020,'' in {\em MAbs}, vol.~12, p.~1703531, Taylor \& Francis, 2020.

\bibitem{nserc2018nserc}
N.~S. and E.~R.~C. of~C. Government~of Canada, ``Nserc strategic network for
  the production of single-type glycoform monoclonal antibodies,'' {\em Natural
  Sciences and Engineering Research Council of Canada (NSERC)}, Jun. 28, 2016.

\bibitem{croughan2015future}
M.~S. Croughan, K.~B. Konstantinov, and C.~Cooney, ``The future of industrial
  bioprocessing: batch or continuous?,'' {\em Biotechnology and
  bioengineering}, vol.~112, no.~4, pp.~648--651, 2015.

\bibitem{rodman2016dynamic}
A.~D. Rodman and D.~I. Gerogiorgis, ``Dynamic simulation and visualisation of
  fermentation: effect of process conditions on beer quality,'' {\em
  IFAC-PapersOnLine}, vol.~49, no.~7, pp.~615--620, 2016.

\bibitem{de1997kinetic}
B.~de~Andr{\'e}s-Toro, J.~Giron-Sierra, C.~Fernandez-Conde, J.~Peinado, and
  F.~Garcia-Ochoa, ``A kinetic model for beer production: simulation under
  industrial operational conditions,'' {\em IFAC Proceedings Volumes}, vol.~30,
  no.~5, pp.~203--208, 1997.

\end{thebibliography}

\clearpage

\clearpage
\begin{appendices}
\section{Environment Details}
\label{sec:Environment Details}

\subsection{ReactorEnv}
\label{sec:appendices CSTR ReactorEnv}
Recall that the reaction is of form A → B. Performing a component balance on reactant A, we obtain the following ordinary differential equation

\begin{equation}
  \frac{dc_A}{dt} =\frac{q_{in}}{\pi r^2h}(c_{Af} - c_A) - k_0 \exp(-\frac{E}{RT})c_A,
\end{equation}
where $c_A$ is the concentration of reactant A in $kmol/m^3$, $t$ is the time in $min$, $q_{in}$ is the volumetric flowrate of the inlet stream in $m^3/min$, $r$ is the radius of the reactor in $m$, $h$ is the level of reaction mixture in the reactor in $m$, $c_{Af}$ is the concentration of reactant A in the feed stream in $kmol/m^3$, $k_0$ is the pre-exponential factor in $min^{-1}$, $E/R$ is the ratio of reaction activation energy to the universal gas constant in $K$ and $T$ is the reaction mixture temperature in $K$.

Similarly, an energy balance can be conducted to obtain the following energy balance equation
\begin{equation}
  \frac{dT}{dt} = \frac{q_{in}}{\pi r^2h}(T_{f} - T) +  \frac{-\Delta H}{\rho c_p} k_0 \exp(-\frac{E}{RT})c_A + \frac{2U}{r\rho c_p}(T_c - T),
\end{equation}
where $T_f$ is the temperature of the feed stream in $K$, $\Delta H$ is the heat of reaction in $kJ/kmol$, $\rho$ is the density of the reaction mixture in $kg/m^3$, $c_p$ is the specific heat capacity of the reaction mixture in $kJ/kg \cdot K$, $U$ is the heat transfer coefficient in $kJ/min \cdot m^2 \cdot K$ and $T_c$ is the coolant temperature.

Finally, deriving an overall material balance around the reactor leads to the following equation
\begin{equation}
  \frac{dh}{dt} = \frac{q_{in}-q_{out}}{\pi r^2},
\end{equation}
where $q_{out}$ is the volumetric flow rate of the contents out of the reactor in $m^3/min$.

A summary of the parameter values used in this project is presented in Table \ref{tb:parameters}.
\begin{table}[hbt!]
  \begin{center}
  \caption{Table of parameter values}\label{tb:parameters}
    \begin{tabular}{ccl}
    Parameter & Unit & Value \\\hline
    $q_{in}$ & $m^3/min$ & $0.1$ \\
    $r$ & $m$ & $0.219$ \\
    $c_{Af}$ & $kmol/m^3$ & $1.0$ \\
    $T_f$ & $K$ & 76.85 \\
    $E/R$ & $K$ & $8750.0$ \\
    $k_0$ & $min^{-1}$ & $7.2 \times 10^{10}$ \\
    $-\Delta H$ & $J/mol$ & $5.0 \times 10^4$ \\
    $U$ & $kJ/min \cdot m^2 \cdot K$ & $5.0 \times 10^4$ \\
    $c_p$ & $kJ/kg \cdot K$ & $0.239$ \\
    $\rho$ & $kg/m^3$ & $1000.0$ \\ \hline
    \end{tabular}
  \end{center}
\end{table}

In the CSTR process model described above, $c_A$, $T$ and $h$ are the state variables. The controlled variables are $c_A$ and $h$ while the manipulated variables are $q_{out}$ and $T_c$.

\subsection{AtropineEnv}
\label{sec:appendices AtropineEnv}

\begin{figure}[!ht]
  \begin{center}
    \includegraphics[width=0.95\textwidth]{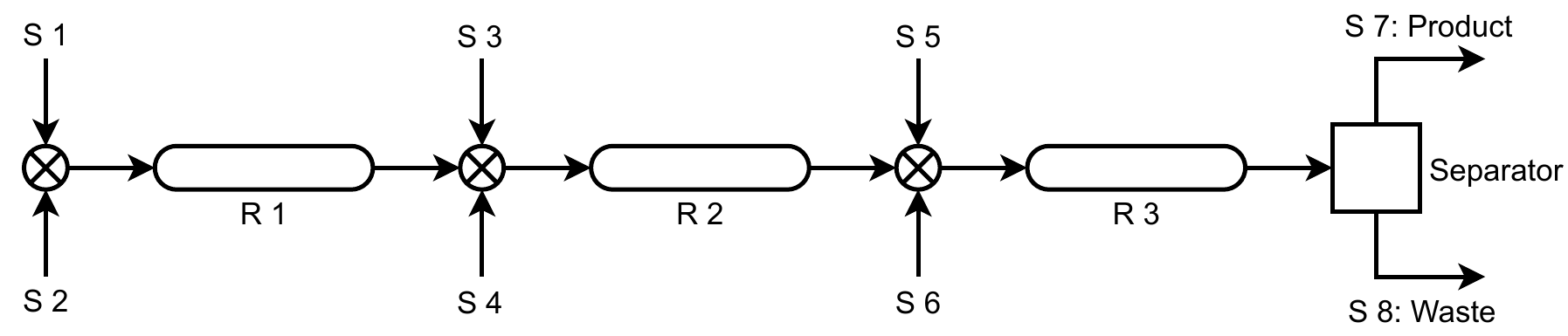}    
    \caption{Process flow diagram of the continuous manufacturing process} 
    \label{fig:appendices process_flow_diagram}
  \end{center}
\end{figure}
A description of the streams in Figure \ref{fig:appendices process_flow_diagram} is summarized in Table \ref{tb:streams}. 
\begin{table}[!ht]
  \begin{center}
  \caption{Description of streams in Figure \ref{fig:appendices process_flow_diagram}} \label{tb:streams}
    \begin{tabular}{cl}
    Stream & Description \\\hline
    S 1 & Tropine in dimethylformamide (2 M)\\
    S 2 & Phenylacetylchloride (pure) \\
    S 3 & Formaldehyde (37 wt\%) \\
    S 4 & Sodium hydroxide (4 M)\\
    S 5 & Buffer solution \\
    S 6 & Organic solvent (Toluene)\\
    S 7 & Product \\ 
    S 8 & Waste \\\hline
    \end{tabular}
  \end{center}
\end{table}

The mixing in the mixers is assumed to occur instantaneously which implies zero dynamics. Thus, the mixer is modeled by the following set of algebraic equations
\begin{equation}\label{eqn:atp_mass_flow}
    \dot{m}_{\text{out},i} = \sum_{k=1}^{n_s} \dot{m}_{\text{in},i,k}
\end{equation}
where $m$ denotes the mass flow rate in xx and $n_s$ is the number of streams. In Equation~\eqref{eqn:atp_mass_flow}, the subscripts $i,k,\text{out},\text{in}$ refer to species, stream number, reactor outlet and reactor inlet respectively.

Each reactor is described by the following partial differential equations obtained from their mass balance 
\begin{equation}\label{eqn:atp_reactor}
    \frac{\partial c_{i,z}}{\partial t} = -Q_{\text{tot}} \frac{\partial c}{\partial V}\bigg|_{i,z} + r_{i,z}
\end{equation}
Equation~\ref{eqn:atp_reactor} can be converted to ordinary differential equations using the method of lines (MOL) \cite{schiesser2012} to obtain
\begin{equation}\label{eqn:atp_reactor_ode}
    \frac{d c_{i,j}}{d t} = -Q_{\text{tot}} \frac{c_{i,j}-c_{i,j-1}}{\Delta V} + r_{i,j}
\end{equation}.
In Equations~\ref{eqn:atp_reactor} and \ref{eqn:atp_reactor_ode}, $Q_{tot}$ is the total volumetric flow rate inside the reactor in $mL/\text{min}$, $r$ is the rate of reaction and $\Delta V$ is the volume of a segment of the reactor. The subscripts $i$ is as previously defined and $j$ is the volume of a segment of the reactor in $mL$.
 The temperature dynamics in each reactor are assumed to be fast and therefore the energy balances are not required. The liquid-liquid separator is described by both ordinary differential equations and algebraic equations which results in a differential-algebraic equation (DAE) system of index 1. More details of the process model can be found in \cite{nikolakopoulou2020} and the references therein. A summary of the key process parameters is shown in Table \ref{tb:process_parameters}. 

\begin{table}[!ht]
  \begin{center}
  \caption{Key process parameters}\label{tb:process_parameters}
    \begin{tabular}{lll}
    Parameter & Description & Value [units] \\\hline
    $V_1$ & Volume of Reactor 1 & 2 [mL]\\
    $V_2$ & Volume of Reactor 2 & 9.5 [mL] \\
    $V_3$ & Volume of reactor 3 & 9.5 [mL]\\
    $V_4$ & Volume of Liquid-liquid separator & 110 [mL]\\ 
    $T_1$ & Temperature of Reactor 1 & 373.15 [K] \\
    $T_2$ & Temperature of Reactor 2 & 373.15 [K] \\
    $T_3$ & Temperature of Reactor 3 & 323.15 [K] \\
    $q_5$ & Volumetric flow rate of $S 5$ & 0.2 [mL/min] \\
    $q_6$ & Volumetric flow rate of $S 6$ & 0.5 [mL/min] \\
    log(D$_9$) & Separation coefficient of atropine & -2 [-] \\\hline
    \end{tabular}
  \end{center}
\end{table}

In the continuous-flow manufacturing process, the volumetric flow rates of streams S1--S4 are manipulated to control the production of atropine while the volumetric flow rates of streams S5 and S6 are kept constant.

\subsubsection{Process control with MPC}

The entire system of DAEs described in the previous section can be written in the form
\begin{align*}
  \dot{x}(t) = f(x(t),z(t),u(t)) \\
  0 = g(x(t),z(t),u(t)) \\
  y(t) = h(x(t), z(t),u(t)),
\end{align*}
where $\dot{x}(t) \in \mathbb{R}^{1694}$ is the velocity of the state vector $x(t) \in \mathbb{R}^{1694}$ at time $t \in \mathbb{R}_+$, $u(t) \in \mathbb{R}^4$ is the vector of inputs, $z(t) \in \mathbb{R}^{30}$ is the vector of algebraic states and $y(t) \in \mathbb{R}$ is the output.

The control objective is to maximize atropine production while minimizing the waste produced. This metric is known as the environmental factor (E-factor) and is defined as
\begin{equation*}
  \text{E-factor} = \frac{\text{Mass of waste produced (excluding water)}}{\text{Mass of product obtained}}.
\end{equation*}

\begin{figure}[!ht]
  \begin{center}
    \includegraphics[width=0.8\textwidth]{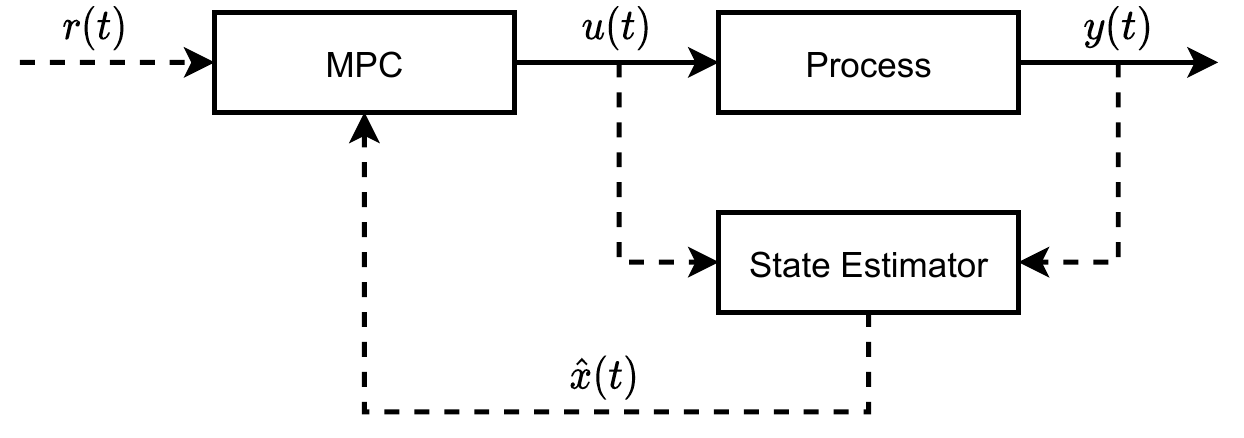}    
    \caption{Schematic diagram of the control system} 
    \label{fig:control_system_diagram}
  \end{center}
\end{figure}

The above DAE system of equations, when used in a model-based controller such as MPC will result in a large-scale nonlinear and possibly non-convex optimization problem which is in general difficult to solve. Thus, to reduce the complexity of the controller, a simple linear model was identified from data and used to make predictions in the controller. A linear discrete-time subspace model relating the inputs to the output (E-factor) was obtained and used in the controller. Since the states of the linear subspace model have no physical meaning, a steady-state Kalman filter was designed to estimate the initial state from the inputs and outputs. A schematic diagram of the control system is shown in Figure \ref{fig:control_system_diagram}. In the \ref{fig:control_system_diagram}, $r(t)$ is the reference signal to be tracked (usually obtained from a higher decision making body such as Real-Time Optimizer (RTO)) and $\hat{x}(t)$ is the initial state estimate for the linear model in the controller. It is worth mentioning that the linear model in the controller may have to be re-identified if the new reference is far from the current reference point.

The parameters for the identified model are

\begin{align*}
  x(k+1) &=
  \begin{bmatrix}
  0.8543 & -0.1164 \\
  0.0195 & 0.8576 
  \end{bmatrix} 
  x(k) +
  \begin{bmatrix}
  -0.0382 & -0.0547 & 0.0103 & 0.1290 \\
  -0.0051 & 0.0072 & 0.0020 & 0.0078 
  \end{bmatrix} 
  u(k) \\
  y(k) &= 
  \begin{bmatrix}
    -148.6124 & -46.8132
  \end{bmatrix} x(k)
\end{align*}
with the associated optimal steady-state Kalman filter gain being
\begin{equation*}
  K = \begin{bmatrix}
    -0.0093 \\
    0.0115
  \end{bmatrix}.
\end{equation*}

A summary of the steady-state values as well as the system constraints are presented in Table \ref{tb:constraints}.


\begin{table}[!ht]
  \begin{center}
  \caption{Summary of the input and output constraints, and their steady-state values}\label{tb:constraints}
    \begin{tabular}{clcc}
    Input & Description & Steady state value [units] & Bounds \\\hline
    $q_1$ & Volumetric flow rate of $S 1$ & 0.4078 [mL] & [0, 5]\\
    $q_2$ & Volumetric flow rate of $S 2$ & 0.1089 [mL] & [0, 5]\\
    $q_3$ & Volumetric flow rate of $S 3$ & 0.3888 [mL] & [0, 5]\\
    $q_4$ & Volumetric flow rate of $S 4$ & 0.2126 [mL] & [0, 5]\\ 
    $y$ & E-factor & 13.057 [kg/kg] & unbounded\\\hline
    \end{tabular}
  \end{center}
\end{table}



\subsection{mAbEnv}
\label{sec:appendices mAbEnv}


\subsubsection{Mathematical model development}
\label{sec:mAb Mathematical model development}
In this section, we present a physics-based mathematical model of the Monoclonal Antibody (mAb) production process. The mAb production process consists of two sub-processes referred to in this work as the upstream and downstream processes. The upstream model presented here is primarily based on the works by Kontoravdi et al. \cite{KONTORAVDI2008,KONTORAVDI2010} as well as other works in literature and the downstream model is mainly based on the works by Gomis-Fons et al. \cite{gomis2020model}. We begin the section by first describing the mAb production process. Subsequently, we present the mathematical models of the various units in the mAb production process.

\begin{figure}[hbt!]
  \begin{center}
    \includegraphics[width=0.9\textwidth]{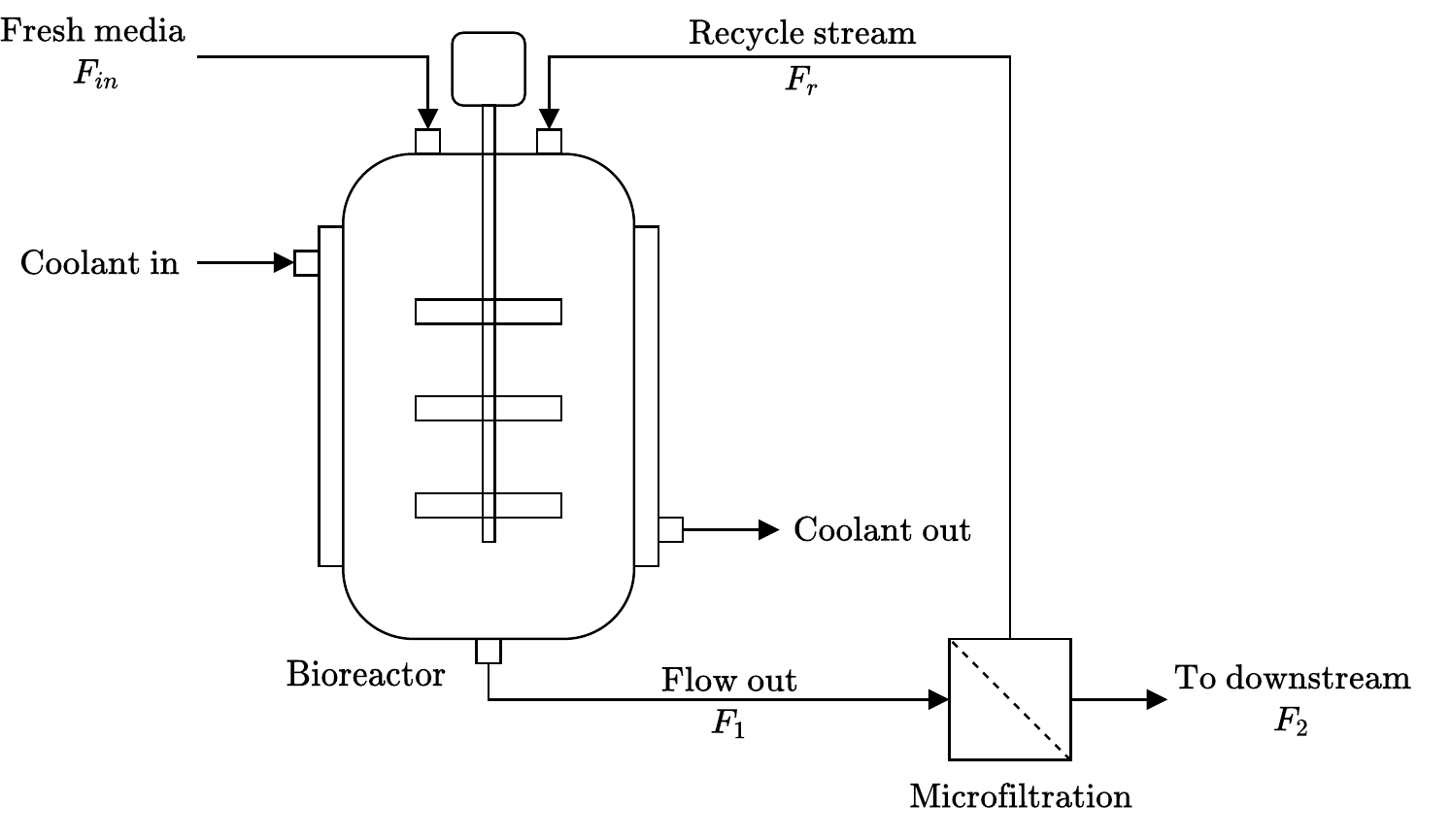}    
    \caption{A schematic diagram of the upstream process for mAb production} 
    \label{fig:upstream}
  \end{center}
\end{figure}

\paragraph{Process description}
As mentioned earlier, the mAb production process consists of the upstream and the downstream processes. In the upstream process, mAb is produced in a bioreactor which provides a conducive environment for mAb growth. The downstream process on the other hand recovers the mAb from the upstream process for storage. In the upstream process for mAb production, fresh media is fed into the bioreactor where a conducive environment is provided for the growth of mAb. A cooling jacket in which a coolant flows is used to control the temperature of the reaction mixture. The contents exiting the bioreactor are passed through a microfiltration unit which recovers part of the fresh media in the stream. The recovered fresh media is recycled back into the bioreactor while the stream with a high amount of mAb is sent to the downstream process for further processing. A schematic diagram of the upstream process is shown in Figure \ref{fig:upstream}. 

\begin{figure}[hbt!]
  \begin{center}
    \includegraphics[width=0.9\textwidth]{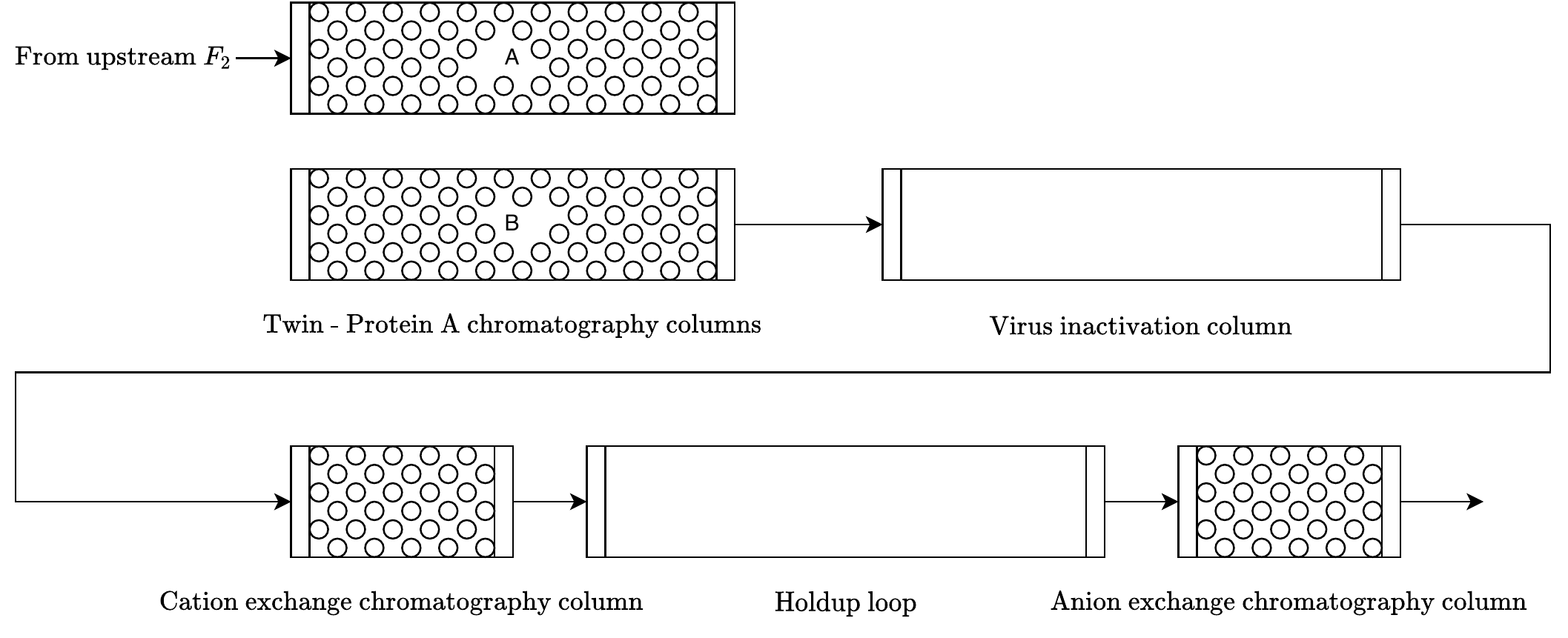}    
    \caption{A schematic diagram of the downstream process for mAb production} 
    \label{fig:downstream}
  \end{center}
\end{figure}

The objective of the downstream process for mAb production is to purify the stream with a high concentration of mAb from the upstream and obtain the desired product. The configuration of the downstream is adopted from Gomis-Fons' work \cite{gomis2020model}. It is composed of a set of fractionating columns, for separating mAb from impurities, and holdup loops, for virus inactivation (VI) and pH conditioning. The schematic diagram of the downstream process is shown in Figure \ref{fig:downstream}. Three main steps are considered in the scheme: capture, virus inactivation, and polish. It is worth mentioning that the ultrafiltration preparing the final product is not considered in this work, and hence is not included in the diagram. The capture step serves as the main component in the downstream and the majority of mAb is recovered in this step. Protein A chromatography columns are usually utilized to achieve this goal. The purpose of VI is to disable the virus and prevent further mAb degradation. At last, the polish step further removes the undesired components caused by VI and cation-exchange chromatography (CEX) and anion-exchange chromatography (AEX) are generally used. In order to obtain a continuous manufacturing process, the perfusion cell culture, a continuous mAb culturing process is used in the upstream, however, the nature of chromatography is discontinuous. Therefore, a twin-column configuration is implemented in the capture step. According to the diagram, column A is connected to the stream from the upstream and loaded with the solutions. Simultaneously, column B is connected to the remaining components of the downstream and conducts further mAb purification. According to Gomis-Fons, et al. \cite{gomis2020model}, the time needed for loading is designed as the same as the time required for the remaining purification steps. Hence, when column A is fully loaded, column B is empty and the resin inside is regenerated. Then, the roles of these two columns will be switched in the new configuration, meaning column B will be connected to the upstream and column A will be connected to the remaining components in downstream. In conclusion, a continuous scheme of downstream is achieved by implementing the twin-column configuration in the capture step. 

\paragraph{Bioreactor modeling}
The mathematical model of the bioreactor can be divided into three parts, namely cell growth and death, cell metabolism, and mAb synthesis and production. Papathanasiou and coworkers described a simplified metabolic network of GN-NS0 cells using a Monod kinetic model \cite{PAPATHANASIOU2017}. 
In the study by Villiger et al. \cite{VILLIGER2016}, while the specific productivity of mAb was observed to be constant with respect to viable cell density, it varied with respect to the extracellular pH. By considering these two models, we proposed one simplified model to describe the continuous upstream process. The following assumptions were used in developing the dynamic model of the bioreactor in the continuous production of the mAb process.

\begin{itemize}
    \item The contents of the bioreactor are perfectly mixed 
    \item The dilution effect is negligible
    \item The enthalpy change due to cell death is negligible
    \item There is no heat loss to the external environment
    \item The temperature of the recycle stream and the temperature of the reaction mixture are equal
\end{itemize}

\subparagraph{Cell growth and death}
An overall material balance on the bioreactor yields the equation
\begin{align} \label{eq:mb:vessel}
    \frac{d{V_1}}{dt} &= F_{in} + F_{r} - F_{out}. 
\end{align}
In Equation~\eqref{eq:mb:vessel}, $V$ is the volume in $L$, and $F_{in}$, $F_{r}$, and $F_{out}$ are the volumetric flow rate of the fresh media into the reactor, the volumetric flow rate of the recycle stream and the volumetric flow rate out of the bioreactor respectively in $L/min$. Throughout this report, the subscripts $1$ and $2$ represent the bioreactor and the microfiltration unit respectively.

The conversion of the viable and total cells within the culture can be described using a component balance on the viable and total number of cells as shown in Equations~\eqref{eq:mb:viablecell:R1} and \eqref{eq:mb:totalcell:R1}
\begin{align}
    \frac{dX_{v1}}{dt} &= \mu X_{v1} - \mu_d X_{v1} - \frac{F_{in}}{V_1}X_{v1}+ \frac{F_{r}}{V_1}(X_{vr}-X_{v1}) \label{eq:mb:viablecell:R1}\\
    \frac{dX_{t1}}{dt} &=  \mu X_{v1} -  \frac{F_{in}}{V_1}X_{t1} + \frac{F_{r}}{V_1}(X_{tr}-X_{t1}), \label{eq:mb:totalcell:R1}
\end{align}
where $X$ is the cell concentration in $cells/L$, $\mu$ is the specific growth rate in $min^{-1}$ and $\mu_d$ is the specific death rate in $min^{-1}$. The subscripts $v$ and $t$ denote the viable and total cells respectively.

The specific cell growth rate is determined by the concentrations of the two key nutrients namely glucose and glutamine, the two main metabolites namely lactate and ammonia and temperature following the Monod kinetics
\begin{align}
    \mu &= \mu_{max}f_{lim}f_{inh} \label{eq:mu}\\
    f_{lim} &= (\frac{[GLC]_1}{K_{glc}+[GLC]_1})(\frac{[GLN]_1}{K_{gln}+[GLN]_1}) \label{eq:flim} \\
    f_{inh} &= (\frac{KI_{lac}}{KI_{lac}+[LAC]_1}) (\frac{KI_{amn}}{KI_{amn}+[AMN]_1}). \label{eq:finh}
\end{align}
In Equation~\eqref{eq:mu}, $\mu_{max}$ is the maximum specific growth rate in $min^{-1}$, $f_{lim}$ and $f_{inh}$ are the nutrient limitation function and the product inhibition function which are described in Equations~\eqref{eq:flim} and \eqref{eq:finh}, respectively. In Equations \eqref{eq:flim} and \eqref{eq:finh}, $[GLC]$, $[GLN]$, $[LAC]$ and $[AMM]$ stand for the concentrations of glucose, glutamine, lactate and ammonia in $mM$, and $K_{glc}$, $K_{gln}$, $KI_{lac}$ and $KI_{amm}$ represent the Monod constant for glucose, glutamine, lactate and ammonia respectively in $mM$.

The specific death rate is determined based on the assumption that cell death is only a function of the concentration of ammonia accumulating in the culture, and is shown as follows:
\begin{equation}\label{eq:mu_d}
    \mu_d = \frac{\mu_{d,max}}{1+(\frac{K_{d,amm}}{[AMM]_1})^n}, ~ n > 1. 
\end{equation}
In Equation~\eqref{eq:mu_d}, $n$ is assumed to be greater than 1 to give a steeper increase of specific death as ammonia concentration increases.

Temperature is a key factor in the maintenance of cell viability and productivity in bioreactors. It is expected that the growth and death of the mAb-producing cells will be affected by temperature. The effect of temperature on the specific growth and death rates is achieved through the maximum specific growth and death rates. In this study, standard linear regression of data available in literature \cite{jimenez2016} was used to obtain a linear relationship between the temperature and the maximum cell growth rate $\mu_{\text{max}}$. 
\begin{equation}\label{eqn:max_specific_growth}
    \mu_{\text{max}} = 0.0016T - 0.0308.
\end{equation}
Similarly, a linear relationship was obtained for the maximum cell death rate as shown in 
\begin{equation}\label{eqn:max_specific_death}
    \mu_{d,\text{max}} = -0.0045T + 0.1682.
\end{equation}
In \eqref{eqn:max_specific_growth} and \eqref{eqn:max_specific_death}, $T$ is the temperature of the bioreactor mixture in $^\circ C$. The data was obtained for the maximum specific growth and death rates at 33 $^{\circ}$C and 37 $^{\circ}$C. Therefore, the Equations \eqref{eqn:max_specific_growth} and \eqref{eqn:max_specific_death} are valid only within this temperature range. A heat balance on the bioreactor together with the following above assumptions leads to the following ordinary differential equation:
\begin{equation}\label{eqn:temp}
    \frac{dT}{dt}=\frac{F_{in}}{V_1}(T_{in}-T) +\frac{-\Delta H}{\rho c_p}(\mu X_{v1}) + \frac{U}{V_1 \rho  c_p}(T_c - T).
\end{equation} 

In Equation~\eqref{eqn:temp}, $T_{in}$ is the temperature of the fresh media in $^{\circ} C$, $\Delta H$ is the heat of reaction due to cell growth in $J/mol$,  $\rho$ is the density of the reaction mixture in $g/L$, $c_p$ is the specific heat capacity of the reaction in $J/(g \circ C)$, $U$  is the overall heat transfer coefficient in $J/(hr ^\circ C)$), and $T_c$ is the temperature of fluid in the jacket in $^{\circ}$C.

The first term of Equation~\eqref{eqn:temp} represents the heat transfer due to the inflow of the feed and the second term represents the heat consumption due to the growth of the cells. The final term describes the external heat transfer to the bioreactor due to the cooling jacket.

\subparagraph{Cell metabolism}
A mass balance on glucose, glutamine, lactate and ammonia around the bioreactor results in the following equations \cite{PAPATHANASIOU2017}:

\begin{align}
    \frac{d[GLC]_1}{dt} & = -Q_{glc}X_{v1} +  \frac{F_{in}}{V_1} ([GLC]_{in} - [GLC]_1) + \frac{F_{r}}{V_1}([GLC]_r-[GLC]_1) \\
    Q_{glc} &= \frac{\mu}{Y_{X,glc}} + m_{glc} \\
    \frac{d[GLN]_1}{dt} &= - Q_{gln}X_{v1} - K_{d,gln}[GLN]_1 +  \frac{F_{in}}{V_1}([GLN]_{in} - [GLN]_1) -  \frac{F_{r}}{V_1}([GLC]_1 - [GLN]_1)\\
    Q_{gln} &= \frac{\mu}{Y_{X,gln}} + m_{gln} \\
    m_{gln} &= \frac{\alpha_1 [GLN]_1}{\alpha_2+[GLN]_1}
\end{align}
\begin{align}
    \frac{d[LAC]_1}{dt} &= Q_{lac}X_{v1} -  \frac{F_{in}}{V_1}[LAC]_1 + \frac{F_r}{V_1}([LAC]_r-[LAC]_1) \\
    Q_{lac} &= Y_{lac,glc}Q_{glc}\\
    \frac{d[AMM]_1}{dt} &= Q_{amm}X_{v1} + K_{d,gln}[GLN]_1 -  \frac{F_{in}}{V_1}[AMM]_1 + \frac{F_r}{V_1}([AMM]_r-[AMM]_1) \\
    Q_{amm} &= Y_{amm,gln}Q_{gln}.
\end{align}

\subparagraph{MAb production}
The rate of mAb production is described as
\begin{align}
    \frac{d[mAb]_1}{dt} &= X_{v1} Q_{mAb} -  \frac{F_{in}}{V_1}[mAb]_1 + \frac{F_r}{V_1}([mAb]_r-[mAb]_1) \label{eq:mAb:R1} \\
    Q_{mAb} &= Q_{mAb}^{max} exp[-\frac{1}{2}(\frac{pH-pH_{opt}}{\omega_{mAb}})^2] \label{eq:QmAb:R1}.
\end{align}
In Equation~\eqref{eq:QmAb:R1}, $Q_{mAb}^{max}$ is the maximum specific productivity with unit $mg/cell/min$, and $\omega_{mAb}$ is the pH-dependent productivity constant. $pH_{opt}$ is the optimal culture pH as shown in \cite{VILLIGER2016}. The pH value is assumed as a function of state and shown in Section \ref{sec:pH}.

\paragraph{Mircofiltration}
\subparagraph{Cell separation}
In the cell separation process, a external hollow fiber (HF) filter is used as cell separation device. It is assumed that no reactions occur in the separation process. Hence, the concentration of each variable in recycle stream is shown as follows:
\begin{align}
    X_{vr} & = \eta_{rec} X_{v1}\frac{F_1}{F_r}\\
    X_{tr} & = \eta_{rec} X_{t1}\frac{F_1}{F_r}\\
    [GLC]_r & = \eta_{ret} [GLC]_1\frac{F_1}{F_r}\\
    [GLN]_r & = \eta_{ret} [GLN]_1\frac{F_1}{F_r}\\
    [LAC]_r & = \eta_{ret} [LAC]_1\frac{F_1}{F_r}\\
    [AMM]_r & = \eta_{ret} [AMM]_1\frac{F_1}{F_r}\\
    [mAb]_r & = \eta_{ret} [mAb]_1\frac{F_1}{F_r}.
\end{align}
According to \cite{clincke2013}, the cell recycle rate ($\eta_{rec}$) is assumed to be 92$\%$ and the retention rates of glucose, glutamine, lactate, ammonia, and mAb ($\eta_{ret}$) are assumed to be 20$\%$.

The material balance around the separation device is shown as:
\begin{equation}
    \frac{dV_2}{dt} = F_1  - F_2 -F_r.
\end{equation}

The mass balance for concentrations of glucose, glutamine, lactate, ammonia, and mAb can be described as:
\begin{align}
    \frac{dX_{v2}}{dt} &= \frac{F_1}{V_2}(X_{v1}-X_{v2}) - \frac{F_r}{V_2} (X_{vr} - X_{v2}) \\
    \frac{dX_{t2}}{dt} &= \frac{F_1}{V_2}(X_{t1}-X_{t2}) - \frac{F_r}{V_2} (X_{tr} - X_{t2}) \\
    \frac{d[GLC]_2}{dt} &= \frac{F_1}{V_2}([GLC]_1-[GLC]_2) - \frac{F_r}{V_2} ([GLC]_r - [GLC]_2) \\
    \frac{d[GLN]_2}{dt} &= \frac{F_1}{V_2}([GLN]_1-[GLN]_2) - \frac{F_r}{V_2} ([GLN]_r - [GLN]_2) \\
    \frac{d[LAC]_2}{dt} &= \frac{F_1}{V_2}([LAC]_1-[LAC]_2) - \frac{F_r}{V_2} ([LAC]_r - [LAC]_2) \\
    \frac{d[AMM]_2}{dt} &= \frac{F_1}{V_2}([AMM]_1-[AMM]_2) - \frac{F_r}{V_2} ([AMM]_r - [AMM]_2) \\
    \frac{d[mAb]_2}{dt} &= \frac{F_1}{V_2}([mAb]_1-[mAb]_2) - \frac{F_r}{V_2} ([mAb]_r - [mAb]_2). 
\end{align}

\subparagraph{pH value} \label{sec:pH}
pH is defined as the decimal logarithm of the reciprocal of the hydrogen ion activity in a solution. We assume our pH model as follows:
\begin{equation}
    pH = \theta_1 - log_{10}(\theta_2 [AMM] +\theta_3).
\end{equation}
After applying nonlinear regression method, we fit the model as:
\begin{equation}
    pH = 7.1697 - log_{10}(0.074028[AMM] +0.968385).
\end{equation}

\paragraph{Downstream modeling}
The mathematical model of the downstream is constructed based on each unit operation. Specifically, two different models are utilized to describe the loading mode and elution mode of the Protein A chromatography column separately. The models for CEX and AEX share the same mathematical structure with different parameters and the models for VI and holdup loop share the same structures and parameters. A detailed explanation of each model is shown in the following subsections.

\begin{figure}[hbt!]
  \begin{center}
    \includegraphics[width=0.6\textwidth]{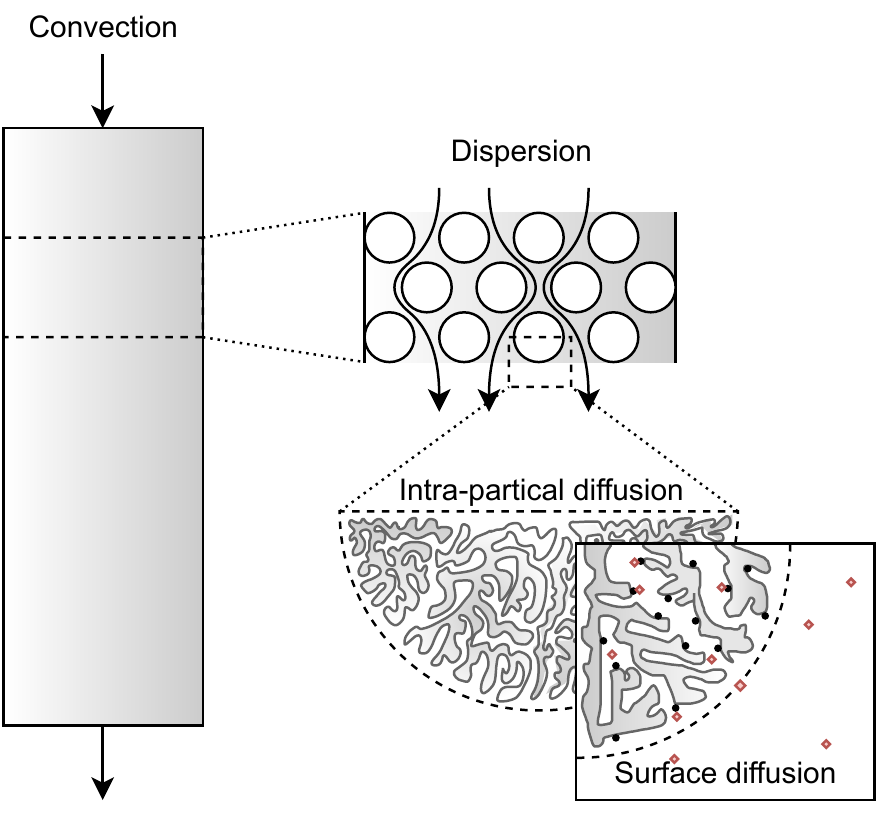}    
    \caption{A schematic diagram of the chromatography column}
    \label{fig:chromatography}
  \end{center}
\end{figure}

\subparagraph{Protein A chromatography column loading mode}
A schematic diagram \cite{dizaji2016minor} depicting a general chromatography column is shown in Figure~\ref{fig:chromatography}. The column is packed with the porous media which have the binding sites with mAb. The porous media is defined as the stationary phase and the fluid which contains mAb and flows through the column is considered as the mobile phase. Three types of mass transfers are usually considered inside of the column. From the top of the figure, the convection caused by the bulk fluid movement is portrayed. Then by only considering a control volume of the column, which is illustrated in the second subfigure, the dispersion of mAb along the axial direction is shown. Within the beads, there is intra-particle diffusion and in the last subfigure, mAbs are adsorbed on the binding sites of beads.

The general rate model (GRM) simulates the mass transfer in a chromatography column, with the assumption that the transfer along the radial direction of the column is negligible and the transfer along the axial direction of the column and the radial direction in the beads is considered. 

In this work, the GRM identified by Perez-Almodovar and Carta \cite{perez2009igg} is used to describe the loading mode of the Protein A chromatography column. The mass transfer along the axial coordinate is described below:

\begin{equation}\label{eq:capture:mobilephase}
    \frac{\partial c}{\partial t} = D_{ax} \frac{\partial^2 c}{\partial z^2} - \frac{v}{\epsilon_c}\frac{\partial c}{\partial z} - \frac{1-\epsilon_c}{\epsilon_c} \frac{3}{r_p} k_f(c-c_p |_{r=r_p}),
\end{equation}
where $c$ denotes the mAb concentration in the mobile phase, changing with time ($t$) and along with the axial coordinates of columns ($z$). $D_{ax}$ is the axial dispersion coefficient, $v$ is the superficial fluid velocity, $\epsilon_c$ is the extra-particle column void, $r_p$ is the radius of particles and $k_f$ is the mass transfer coefficient.

On the right-hand side of Equation~\eqref{eq:capture:mobilephase}, there are three terms. The first term, $\frac{\partial^2 c}{\partial z^2}$, models the dispersion of mAb. In other words, it describes the movement of mAb caused by the concentration difference in the column. The second term $\frac{\partial c}{\partial z}$ denotes the change of concentration of mAb caused by the convection flow. The last term $k_f(c-c_p |_{r=r_p})$ describes the mass transfer between the mobile phase $c$ and the surface of the beads $c_p |_{r=r_p}$.

The boundary conditions of Equation~\eqref{eq:capture:mobilephase} are shown below:
\begin{subequations}
\begin{align}
    \frac{\partial c}{\partial z} &= \frac{v}{\epsilon_c D_{ax}} (c-c_F) \mbox{ ~~at~~} z=0 \label{eq:capture:mobilephase:boundary1}\\
    \frac{\partial c}{\partial z} &= 0 \mbox{ ~~at~~} z=L, \label{eq:capture:mobilephase:boundary2}
\end{align}
\end{subequations}
where $c_F$ stands for the harvest mAb concentration from the upstream process. 

The concentration of mAb along radial coordinate in the beads ($c_p$) is the second component of GRM and the mass balance for protein diffusion inside the porous particles is shown in Equation~\eqref{eq:capture:particle} with boundary conditions in Equations~\eqref{eq:capture:particle:boundary1} and~\eqref{eq:capture:particle:boundary2}

\begin{equation}\label{eq:capture:particle}
     \frac{\partial c_p}{\partial t} = D_{eff} \frac{1}{r^2} \frac{\partial}{\partial r}(r^2 \frac{\partial c_p}{\partial r}) - \frac{1}{\epsilon_p} \frac{\partial (q_1 + q_2)}{\partial t} 
\end{equation}
\begin{subequations}
\begin{align}
    \frac{\partial c_p}{\partial r} &= 0 \mbox{ ~~at~~} r=0 \label{eq:capture:particle:boundary1}\\
    \frac{\partial c_p}{\partial r} &= \frac{k_f}{D_{eff}} (c-cp) \mbox{ ~~at~~} r=r_p, \label{eq:capture:particle:boundary2}
\end{align}
\end{subequations}
where $D_{eff}$ is the effective pore diffusivity, $r$ is the distance from the current location to the center of the particle, and $\epsilon_p$ is the particle porosity.

At last, the description of adsorbed mAb concentration ($q_1$ and $q_2$) is shown as follows:
\begin{equation}\label{eq:capture:adsobed}
   \frac{\partial q_i}{\partial t} = k_i [(q_{max,i}-q_i)c_p|_{r=r_p} - \frac{q_i}{K}] \mbox{ ~~for~~} i=1,2,
\end{equation}
where $k_i$ is the adsorption kinetic constant, $q_{max}$ is the column capacity, and $K$ is the Langmuir equilibrium constant. The reason for having two $\frac{\partial q}{\partial t}$ is because there are two adsorption sites on the beads and one of them is a fast binding site and another one is the slow one.

\subparagraph{Protein A chromatography column elution mode}
An adsorption kinetic model, convective-dispersive equation with adsorption, is used to describe the elution of the Protein A chromatography column. The setup of boundary conditions for this model can take Equations~\eqref{eq:capture:mobilephase:boundary1} and \eqref{eq:capture:mobilephase:boundary2} as the reference, at the same time keeping the inlet and outlet conditions of elution mode in mind. The model is shown as follows:
\begin{align}
    \frac{\partial c}{\partial t} &= D_{ax} \frac{\partial^2 c}{\partial z^2} - \frac{v}{\epsilon}\frac{\partial c}{\partial z} + \frac{1-\epsilon_c}{\epsilon} \frac{\partial q}{\partial t} \label{eq:elution:mobilephase} \\
    \frac{\partial q}{\partial t} &= k [H_0 c_s^{-\beta} (1- \frac{q}{q_{max}})c-q] \label{eq:elution:adsorbed}\\
    \frac{\partial c_s}{\partial t} &=D_{ax}\frac{\partial^2c_s}{\partial z^2} - \frac{v}{\epsilon} \frac{\partial c_s}{\partial z}, \label{eq:elution:modifier}
\end{align}
where $c$ is the mAb concentration in the mobile phase, $c_s$ stands for the modifier concentration, $q$ is the adsorbed mAb concentration. $k$ is the adsorption/desorption rate, $H_0$ is the Henry equilibrium constant, $\beta$ is the equilibrium modifier-dependence parameter, and $\epsilon$ is the total column void.

On the right hand side of Equation~\eqref{eq:elution:mobilephase}, the first two terms are similar with those in Equation~\eqref{eq:capture:mobilephase}. The third term $\frac{\partial q}{\partial t}$ is detailed expressed in Equation~\eqref{eq:elution:adsorbed}, which is a Langmuir isotherm describing the adsorption and desorption of mAb on beads. This mass transfer is affected by the concentration of the modifier $c_s$ whose dynamics are described in Equation~\eqref{eq:elution:modifier}.

\subparagraph{CEX and AEX chromatography}
The adsorption kinetic model shown in Equations~\eqref{eq:elution:mobilephase},~\eqref{eq:elution:adsorbed} and Equation~\eqref{eq:elution:modifier} can also be used to describe the CEX and AEX chromatography process. The same rule applies to the boundary conditions.
Since the AEX column is in flow-through mode as described in \cite{perez2009igg}, the product mAb is not adsorbed on the beads and the kinetic constant $k$ is supposed to be zero.

\subparagraph{Virus inactivation and holdup pool}
Equation~\eqref{eq:loops} shows the model of loop for VI and holdup, which is modeled as a one-dimensional dispersive-convective transport, with boundary conditions in Equations~\eqref{eq:capture:mobilephase:boundary1} and~\eqref{eq:capture:mobilephase:boundary2}. Since the loop is not packed, there is no intra-particle diffusion or mass transfer between mAb outside of particles and on the surface of the particles.

\begin{equation} \label{eq:loops}
    \frac{\partial c}{\partial t} = D_{ax} \frac{\partial^2 c}{\partial z^2} - v\frac{\partial c}{\partial z}. 
\end{equation}

\subsubsection{Control problem formulation and controller design}

In this chapter, we present preliminary results of implementing advanced process control (APC) techniques in the operation of the continuous mAb production process. Specifically, two variants of APC algorithms, namely model predictive control (MPC) and economic model predictive control (EMPC) were designed and tested on the mAb production process. We begin the chapter by presenting the control problem to be addressed. Subsequently, we present the various controller designs. Finally, we compare the results of MPC and EMPC.

\paragraph{Control problem formulation}

\subparagraph{Upstream process}

Before we begin this section, let us rewrite the model of the upstream mAb production process in the state space form
\begin{equation}\label{eqn:state_space}
    \dot{x}(t) = f(x(t),u(t)),
\end{equation}
where $\dot{x}(t) \in \m R^{15}$ is the velocity of the state vector $x \in \m R^{15}$ at time $t$ and $u(t)  \in \m R^{7}$ is the input vector. The variables in the input vector will be defined later in this section. For practical reasons, we assume that the state and input are constrained to be in the spaces $\m X$ and $\m U$ respectively.

The primary control objective in this work is to ensure that safety and environmental regulations are adhered to during the operation of the mAb production process. From an economic point of view, it is essential to maximize the production of mAb in the upstream process. Thus, two secondary economic objectives are considered. The first is the maximization of the mAb flow rate from the bioreactor while the second is the maximization of the mAb flow rate in the separator (microfiltration unit). These objectives are given as
\begin{equation}
    \ell_{\text{bioreactor}} = \text{mAb concentration in bioreactor} \times \text{flow out of the bioreactor}
\end{equation}
\begin{equation}
    \ell_{\text{separator}} = \text{mAb concentration in separator} \times \text{flow out of the separator}.
\end{equation}
Combining the two economic objectives, the following economic objective is obtained:
\begin{equation}
    \ell_e(x,u) = \ell_{\text{bioreactor}} + \ell_{\text{separator}}.
\end{equation}

To achieve these objectives, we manipulate (as input variables) the flow rates $F_{in}$, $F_r$, $F_1$ and $F_2$, the coolant temperature $T_c$ together with the concentration of ammonia and glucose in the fresh media stream. Considering the objectives, advanced process control (APC) algorithms that consider the complex system interaction while ensuring constraint satisfaction must be used. 

Let us define the steady-state economic optimization with respect to the economic objective $\ell_e$ as
\begin{subequations} \label{eqn:ss_opt}
    \begin{align}
        (x_s,u_s) &= \arg \min ~ -\ell_e(x,u) \label{eqn:ss_opt_cost}\\
        \mbox{subject to} ~~~ & 0 = f(x,u) \label{eqn:ss_opt_model}\\ 
        & x \in \m{X} \label{eqn:ss_opt_state_con}\\
        & u \in \m{U}, \label{eqn:ss_opt_input_con}
    \end{align}
\end{subequations}
where Equation~\eqref{eqn:ss_opt_model} is the system model defined in Equation~\eqref{eqn:state_space} with zero state velocity, and Equations~\eqref{eqn:ss_opt_state_con} and~ \eqref{eqn:ss_opt_input_con} are the the constraints on the state and the input respectively. The negative economic cost function converts the maximization problem to a minimization problem. The optimal value function  in \eqref{eqn:ss_opt} is used as the setpoint for MPC to track.

\paragraph{Controller design}

\subparagraph{Tracking Model Predictive Control (MPC)} \label{sec:control_mpc}
MPC is a multivariable advanced process control algorithm which has gained significant attention in the process control community. This is because of its ability to handle the complex system interactions and constraints in the controller design. At each sampling time $t_k$, the following dynamic optimization problem is solved:
\begin{subequations} \label{eqn:mpc_opt}
    \begin{align} 
        \min_{\bf{u}} & ~~~ \int_{t_k}^{t_k + N\Delta} (x(t)-x_s)^T Q (x(t)-x_s) + (u(t)-u_s)^T R (u(t)-u_s) dt \label{eqn:mpc_opt_a}\\
        \mbox{subject to} & ~~~ \dot{x}(t) = f(x(t),v(t)) \label{eqn:mpc_opt_b} \\
        & ~~~ x(t_{k}) = x(t_{k}) \label{eqn:mpc_opt_c}\\
        & ~~~  x(t) \in \m{X} \label{eqn:mpc_opt_d}\\
        & ~~~ u(t) \in \m{U}.  \label{eqn:mpc_opt_e}
    \end{align}
\end{subequations}
In the optimization problem \eqref{eqn:mpc_opt} above, Equation~\eqref{eqn:mpc_opt_b} is the model constraint which is used to make predictions into the future,  Equation~\eqref{eqn:mpc_opt_c} is the initial state constraint, $\Delta$ is the sampling time, $N$ is the prediction and control horizons, Equations~\eqref{eqn:mpc_opt_d} and~\eqref{eqn:mpc_opt_e} are the constraints on the state and input respectively, and $Q$ and $R$ are matrices of appropriate dimensions which represent the weights on the deviation of states and the inputs from the setpoint. The setpoint is obtained by solving the steady-state optimization problem in~\eqref{eqn:empc_opt}. The decision variable \textbf{u} in~\eqref{eqn:mpc_opt} is the optimal input sequence for the process. The first input $u(t_k)$ is applied to the system and the optimization problem is solved again after one sampling time.

\subparagraph{Economic Model Predictive Control (EMPC)}
The MPC described in Section \ref{sec:control_mpc} uses a quadratic cost in its formulation. However, in recent years MPC with a general objective is known as economic MPC (EMPC) has received significant attention. 
The objective function in an EMPC generally reflects some economic performance criterion such as profit maximization or heat minimization. This is in contrast with the tracking MPC described earlier where the objective is a positive definite quadratic function. The integration of process economics directly in the control layer makes EMPC of interest in many areas, especially in the process industry. There have been numerous applications of EMPC. 

At each sampling time $t_k$, the following optimization problem is solved
\begin{subequations} \label{eqn:empc_opt}
    \begin{align} 
        \min_{\bf{u}} & ~~~ \int_{t_k}^{t_k + N\Delta} -\ell_e(x(t),u(t)) dt \label{eqn:empc_opt_a}\\
        \mbox{subject to} & ~~~ \dot{x}(t) = f(x(t),u(t)) \label{eqn:empc_opt_b} \\
        & ~~~ x(t_{k}) = x(t_{k}) \label{eqn:empc_opt_c}\\
        & ~~~  x(t) \in \m{X} \label{eqn:empc_opt_d}\\
        & ~~~ u(t) \in \m{U}.  \label{eqn:empc_opt_e}
    \end{align}
\end{subequations}
In the optimization problem (\ref{eqn:empc_opt}) above, the constraints are the same as the optimization problem in (\ref{eqn:mpc_opt}). However, a general cost function is used in place of the quadratic cost function. The benefits of EMPC over MPC will be demonstrated in the results section.

\paragraph{Simulation settings}

After conducting extensive open-loop tests, the control and prediction horizons $N$ for both controllers were fixed at 100. This implies that at a sampling time of 1 hour, the controllers plan 100 hours into the future. The weights on the deviation of the states and input from the setpoint were identifying matrices. As mentioned earlier, the setpoint for the tracking MPC was determined by solving the optimization problem in \eqref{eqn:ss_opt}. The optimization problems were implemented using the modeling environment casadi \cite{andersson2019} in Python.

\subsubsection{Model parameters}

\begin{table}[hp]
  \begin{center}
  \caption{Parameters for the upstream process model}\label{tb:upstream_parameters}
  \vspace{2mm}
    \begin{tabular}{lll}
        \hline
    Parameter & Unit & Value \\  
        \hline
    $K_{d,amm}$ & $mM$ & $1.76$ \\ 
    $K_{d,gln}$ & $min^{-1}$ & $0.00016$ \\ 
    $K_{glc}$ & $mM$ & $0.75$ \\
    $K_{gln}$ & $mM$ & $0.038$ \\
    $KI_{amm}$ & $mM$ & $28.48$ \\
    $KI_{lac}$  & $mM$ & $171.76$ \\
    $m_{glc}$ & $mmol/(cell \cdot min)$ & $8.2 \times 10^{-16}$\\
    $Q_{mAb}^{max}$ & $mg/(cell\cdot min)$ & $1.1 \times 10^{-11}$\\
    $Y_{amm,gln}$ & $mmol/mmol$ & $0.45$ \\
    $Y_{lac,glc}$ & $mmol/mmol$ & $2.0$ \\
    $Y_{X,glc}$ & $cell/mmol$ & $2.6 \times 10^8$\\
    $Y_{X,gln}$ & $cell/mmol$ & $8.0 \times 10^8$\\
    $\alpha_1$ & $(mM \cdot L)/(cell \cdot min)$ & $5.7 \times 10^{-15}$\\
    $\alpha_2$ & $mM$ & 4.0\\
    $-\Delta H$ & $J/mol$ & $5.0 \times 10^5$ \\
    $rho$ & $g/L$ & $1560.0$\\
    $c_p$ & $J/(g ^\circ C)$ & $1.244$ \\
    $U$ & $J/(h ^\circ C)$ & $4 \times 10^2$ \\
    $T_{in}$ & $^\circ C$ & $37.0$ \\
    
        \hline
    \end{tabular}
  \end{center}
\end{table}

\begin{table}[hp]
  \begin{center}
  \caption{Parameters of digital twin of downstream}\label{tb:para_down}
  \vspace{2mm}
    \begin{tabular}{llll}
        \hline
    Step & Parameter & Unit & Value\\ 
    \hline
    Capture & $q_{max,1}$ & $mg/mL$ & $36.45$\\
     & $k_{1}$ & $mL/(mg~min)$ & $0.704$\\
     & $q_{max,2}$ & $mg/mL$ & $77.85$\\
     & $k_{2}$ & $mL/(mg~min)$ & $2.1\cdot10^{-2}$\\
     & $K$ & $mL/mg$ & $15.3$\\
     & $D_{eff}$ & $cm^{2}/min$ & $7.6\cdot10^{-5}$\\
     & $D_{ax}$ & $cm^{2}/min$ & $5.5\cdot10^{-1}v$\\
     & $k_{f}$ & $cm/min$ & $6.7\cdot10^{-2}v^{0.58}$\\
     & $r_{p}$ & $cm$ & $4.25\cdot10^{-3}$\\
     & $L$ & $cm$ & $20$\\
     & $V$ & $mL$ & $10^5$\\
     & $\epsilon_c$ & $-$ & $0.31$\\
     & $\epsilon_p$ & $-$ & $0.94$\\
     & $q_{max,elu}$ & $mg/mL$ & $114.3$\\
     & $k_{elu}$ & $min^{-1}$ & $0.64$\\
     & $H_{0,elu}$ & $M^{\beta}$ & $2.2\cdot10^{-2}$\\
     & $\beta_{elu}$ & $-$ & $0.2$\\
     \hline
    Loop & $D_{ax}$ & $cm^{2}/min$ & $2.9\cdot10^{2}v$\\
     & $L$ & $cm$ & $600$\\
     & $V$ & $mL$ & $5\cdot10^5$\\
     \hline
    CEX & $q_{max}$ & $mg/mL$ & $150.2$\\
     & $k$ & $min^{-1}$ & $0.99$\\
     & $H_{0}$ & $M^{\beta}$ & $6.9\cdot10^{-4}$\\
     & $\beta$ & $-$ & $8.5$\\
     & $D_{app}$ & $cm^{2}/min$ & $1.1\cdot10^{-1}v$\\
     & $L$ & $cm$ & $10$\\
     & $V$ & $mL$ & $5\cdot10^{4}$\\
     & $\epsilon_{c}$ & $-$ & $0.34$\\
    \hline
    AEX & $D_{app}$ & $cm^{2}/min$ & $1.6\cdot10^{-1}v$\\
     & $k$ & $min^{-1}$ & $0$\\
     & $L$ & $cm$ & 10\\
     & $V$ & $mL$ & $5\cdot10^{4}$\\
     & $\epsilon_{c}$ & $-$ & $0.34$\\
    \hline
    \end{tabular}
  \end{center}
\end{table}

The parameters of the downstream model are obtained from the work of Gomis-Fons et al. \cite{gomis2020model} and several parameters are modified because the process is upscaled from lab scale to industrial scale. They are summarized in Table~\ref{tb:para_down}.

\subsubsection{Market value and importance}
\label{sec:mAb Market Value and Importance}
Drugs based on monoclonal antibodies (mAbs) play an indispensable role in the biopharmaceutical industry in aspects of therapeutic and market potential. In therapy and diagnosis applications, mAbs are widely used for the treatment of autoimmune diseases, cancer, etc. According to a recent publication, mAbs also show promising results in the treatment of COVID-19 \cite{wang2020human}. Until September 22, 2020, 94 therapeutic mAbs have been approved by U.S. Food \& Drug Administration (FDA) \cite{antibody2020antibody} and the number of mAbs approved within 2010-2020 is three times more than those approved before 2010 \cite{kaplon2020antibodies}. In terms of its market value, it is expected to reach a value of \$198.2 billion in 2023. Thus, with the fact that Canada is an active and competitive contributor to the development of high capacity mAb manufacturing processes \cite{nserc2018nserc}, increasing the production capacity of mAb manufacturing processes is immediately necessary due to the explosive growth in the mAb market. Integrated continuous manufacturing of mAbs represents the state-of-the-art in mAb manufacturing and has attracted a lot of attention because of the steady-state operations, high volumetric productivity, and reduced equipment size and capital cost, etc. \cite{croughan2015future}.

\subsection{PenSimEnv}
\label{sec:appendices_PenSimEnv}

In this section, we briefly describe the mathematical model of the PenSimEnv as presented in Goldrick and coworkers \cite{goldrick2015development}. The model of the fermenter consists of the growth, production, morphology and generation of the biomass. The following key equations describe the four regions in the fermenter, namely growing regions, non-growing regions, degenerated regions and autolysed regions.

\begin{equation}\label{eqn:growing_regions}
    \frac{dA_0}{dt} = r_b - r_{diff} - \frac{F_{in}A_0}{V}
\end{equation}

\begin{equation}\label{eqn:non_growing_regions}
    \frac{dA_1}{dt} = r_e - r_b + r_{diff} - r_{deg}\frac{F_{in}A_1}{V}
\end{equation}

\begin{equation}\label{eqn:degenerated_regions}
    \frac{dA_3}{dt} = r_{deg} - r_a - \frac{F_{in}A_3}{V}
\end{equation}

\begin{equation}\label{eqn:autolysed_regions}
    \frac{dA_4}{dt} = r_a - \frac{F_{in}A_4}{V}
\end{equation}
Equations~\eqref{eqn:growing_regions}~--~\eqref{eqn:autolysed_regions} describe the four regions respectively. In these equations, $r_b$ is the rate of branching, $r_{diff}$ denotes the rate of differentiation, $r_e$ is the rate of extension, $r_{deg}$ is the rate of degeneration, $r_a$ is the rate of autolysis, $P$ is the rate of product formation, $h$ is the rate of hydrolysis, $r_m$ is the rate of maintenance, $A_i$ where $i=0,1,2,3,4$ refers to the actively growing regions, non-growing regions, degenerated regions formed through vocuolation and autolysed regions, t is the batch time, $F_{in}$ refers to all the inputs to the process and $V$ is the volume of the fermenter. The total biomass in the system is given as $\sum_{i=0}^4 A_i$.

The product formation, substrate consumption, and the volume of the fermentation mixture is described in Equations~\ref{eqn:product_formation}~--~\ref{eqn:volume}:
\begin{equation}\label{eqn:product_formation}
    \frac{dP}{dt} = r_p - r_h - \frac{F_{in}P}{V}
\end{equation}
\begin{equation}\label{eqn:substrate_consumption}
    \frac{ds}{dt} = -Y_{s/X}r_e - Y_{s/X}r_b - m_s r_m - Y_{s/P}r_P + \frac{F_{s}c_s}{V} + \frac{F_{oil}c_{oil}}{V}
\end{equation}
\begin{equation}\label{eqn:volume}
    \frac{dV}{dt} = F_s + F_{oil} + F_{PAA} + F_a + F_b + F_w - F_{evp} - F_{dis}
\end{equation}
where $s$ is a combined oil and sugar as a single substrate, $Y_{s/X}$ and $Y_{s/P}$ denotes the substrate yield coefficients of biomass and penicillin respectively, $m_s$ refers to the substrate maintenance coefficient, $F_{oil}$ and $c_{oil}$ denotes the feed flow rate and concentration of soya bean oil respectively, $F_s$ and $c_s$ denotes the feed flow rate and concentration of sugar respectively, $F_{PAA}$ is refers to the flow rate of phenylacetic acid, $F_a$ and $F_b$ refers to the to flow rates of the acid and base respectively, $F_w$ is the flow rate of injection water, $F_{evap}$ is the rate of evaporation of the fermenter and $F_{dis}$ is the rate of discharge from the fermenter during production.

Several other equations such as the component balance on the oxygen and nitrogen in the fermenter is also present in the model. A more detailed description can be found in \cite{goldrick2015development}.

\subsection{BeerFMTEnv}
\label{sec:appendices_BeerFMTEnv}

The fermentation unit is a critical component in the beer manufacturing process. The dynamic model of the beer fermentation process, as presented in the work by Rodman et al. \cite{rodman2016dynamic} and de Andres-Toro et al. \cite{de1997kinetic} is described by 7 ordinary differential equations and several temperature dependent parameters. The equations are derived based on the component balances 
\begin{equation}\label{eqn:active_cells}
    \frac{d [X_A]}{dt} = \mu_x [X_A] - \mu_{DT} [X_A] + \mu_L [X_L]
\end{equation}

\begin{equation}
    \frac{d [X_L]}{dt} = - \mu_L [X_L]
\end{equation}

\begin{equation}
    \frac{d [X_D]}{dt} = \mu_{SD} [X_D] + \mu_{DT} [X_A]
\end{equation}

\begin{equation}
    \frac{d[S]}{dt} = \mu_S [X_A]
\end{equation}

\begin{equation}
    \frac{d [EtOH]}{dt} = f \mu_{EtOH} [X_A]
\end{equation}

\begin{equation}
    \frac{d [DY]}{dt} = \mu_{DY} [S] [X_A] - \mu_{AB} [DY] [EtOH]
\end{equation}

\begin{equation}\label{eqn:ethylacetate}
    \frac{d [EA]}{dt} = Y_{EA} \mu_X [X_A]
\end{equation}
In Equations~\ref{eqn:active_cells}~--~\ref{eqn:ethylacetate}, the symbol $[\cdot]$ represents the concentration of a component, $X_A$ denotes the active cells, $X_L$ denotes the latent cells, $X_D$ refers to the dead cells, $S$ represents sugar, $EtOH$ denotes ethanol, $DY$ denotes diacetyls and $EA$ represents ethyl acetate. The parameter $\mu$ denotes the rates, $f$ is the inhibition factor. More details about the parameters can be found in \cite{rodman2016dynamic}.

\section{Compute}
\label{sec:Compute}
The compute is an RTX 3090 GPU, an RTX 2070 GPU and a GTX 1080 GPU with i9-12900k CPU for a total of 5380 GPU hours.

\end{appendices}

\end{document}